%% file: main_yxz_Camera_Ready.tex
\documentclass[10pt,twocolumn,letterpaper]{article}

\usepackage{cvpr}              

\input{preamble}

\definecolor{cvprblue}{rgb}{0.21,0.49,0.74}
\usepackage[pagebackref,breaklinks,colorlinks,allcolors=cvprblue]{hyperref}

\def\paperID{3025} 
\def\confName{CVPR}
\def\confYear{2026}

\title
{
\vspace{-0.1cm}
Interpretable Cross-Domain Few-Shot Learning with Rectified Target-Domain
Local Alignment
\vspace{-0.1cm}
}

\author
{
Yaze Zhao, \quad Yixiong Zou\thanks{Corresponding author.},\quad Yuhua Li,\quad Ruixuan Li  \\
School of Computer Science and Technology, Huazhong University of Science and Technology \\
{\tt\small \{zyaz, yixiongz, idcliyuhua, rxli\}@hust.edu.cn}
}

\begin{document}
\maketitle
\vspace{-0.4cm}
\input{sec/0_abstract}    
\vspace{-0.45cm}
\input{sec/1_intro}

\input{sec/2_Related}

\input{sec/3_method}
\input{sec/4_exp}
{
    \small
    \bibliographystyle{ieeenat_fullname}
    \bibliography{main}
}

\input{sec/X_suppl}

\end{document}

%% file: preamble.tex
\usepackage{pifont}
\usepackage{multirow}
\usepackage{siunitx}
\usepackage{booktabs}
\usepackage{tabularx}
\usepackage{makecell}
\usepackage{hhline}

\usepackage{float}                      
\usepackage{graphicx}
\usepackage{xcolor}
\usepackage{colortbl}
\usepackage{adjustbox}
\usepackage{threeparttable}

%% file: sec/0_abstract.tex
\begin{abstract}
Cross-Domain Few-Shot Learning (CDFSL) adapts models trained with large-scale general data (source domain) to downstream target domains with only scarce training data, where the research on vision-language models (e.g., CLIP) is still in the early stages. Typical downstream domains, such as medical diagnosis, require fine-grained visual cues for interpretable recognition, but we find that current fine-tuned CLIP models can hardly focus on these cues, albeit they can roughly focus on important regions in source domains. Although current works have demonstrated CLIP's shortcomings in capturing local subtle patterns, in this paper, we find that \textbf{the domain gap and scarce training data further exacerbate such shortcomings, much more than that of holistic patterns}, which we call the local misalignment problem in CLIP-based CDFSL. To address this problem, due to the lack of supervision in aligning local visual features and text semantics, we turn to self-supervision information. Inspired by the translation task, we propose the CC-CDFSL method with cycle consistency, which translates local visual features into text features and then translates them back into visual features (and vice versa), and constrains the original features close to the translated back features.
To reduce the noise imported by richer information in the visual modality, we further propose a Semantic Anchor mechanism, which first augments visual features to provide a larger corpus for the text-to-image mapping, and then shrinks the image features to filter out irrelevant image-to-text mapping. 
Extensive experiments on 
various benchmarks, backbones, and fine-tuning methods
show we can (1) effectively improve the local vision-language alignment, (2) enhance the interpretability of learned patterns and model decisions by visualizing patches, and (3) achieve state-of-the-art performance. 
Code is available at 
\href{https://github.com/z-yaz/CC-CDFSL}{CC-CDFSL}. 
\end{abstract}
\vspace{-0.1cm}

%% file: sec/1_intro.tex
\section{Introduction}
\label{sec:intro}

\begin{figure}[ht]  
\centering  
\includegraphics[width=0.95\columnwidth]{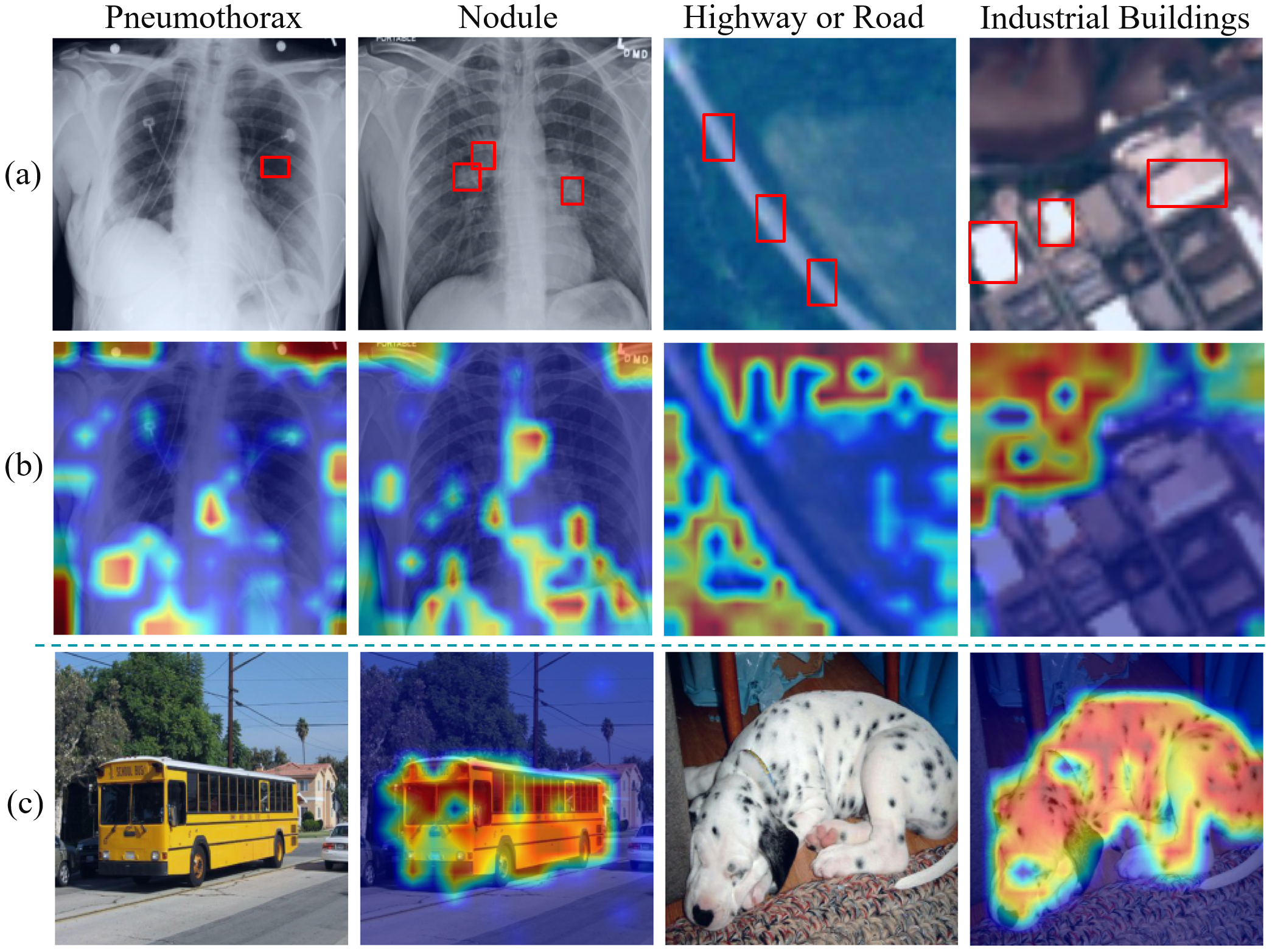}  \vspace{-0.2cm} 
\caption{Fine-grained visual cues (marked as red boxes) are crucial in specialized fields like interpretable medical diagnosis (a). 
However, in target domains, the fine-tuned CLIP cannot focus on these subtle patterns (b), but in source domains (c), CLIP can still roughly capture all important regions for recognition. Therefore, we hypothesize that 
the domain gap and scarce training data exacerbate CLIP's shortcomings in capturing subtle patterns, much more than that of holistic patterns, which we aim to address. 
} 
\label{attnmap} 
\vspace{-0.5cm}  
\end{figure} 

Deep learning has achieved breakthrough progress in computer vision~\cite{DBLP:conf/nips/ChenWDH023, dosovitskiy2021an, zhang2024learning, zhang2024micm, jiang2025revisiting}.
However, these achievements are mainly in general domains where large-scale training data is available. In real-world applications, particularly in downstream domains like the medical diagnosis of rare diseases, acquiring sufficient labeled samples is not only costly but also often constrained by privacy and security concerns. To address this practical challenge, Cross-Domain Few-Shot Learning (CDFSL)~\cite{guo2020broader,zou2024a} has emerged. Its objective is to adapt models from a data-rich source domain to downstream target domains with few labeled data. 
Recently, vision-language foundation models like CLIP~\cite{radford2021learningtransferablevisualmodels}, pretrained on vast image-text pairs, have provided a robust foundation~\cite{DBLP:journals/corr/abs-2111-11432, DBLP:journals/corr/abs-2505-05071} 
for CDFSL due to their generalizability. However, the research is still in its early stages~\cite{zhang2026mind}. 

\begin{figure}[ht]
\centering
\includegraphics[width=0.99\columnwidth]
{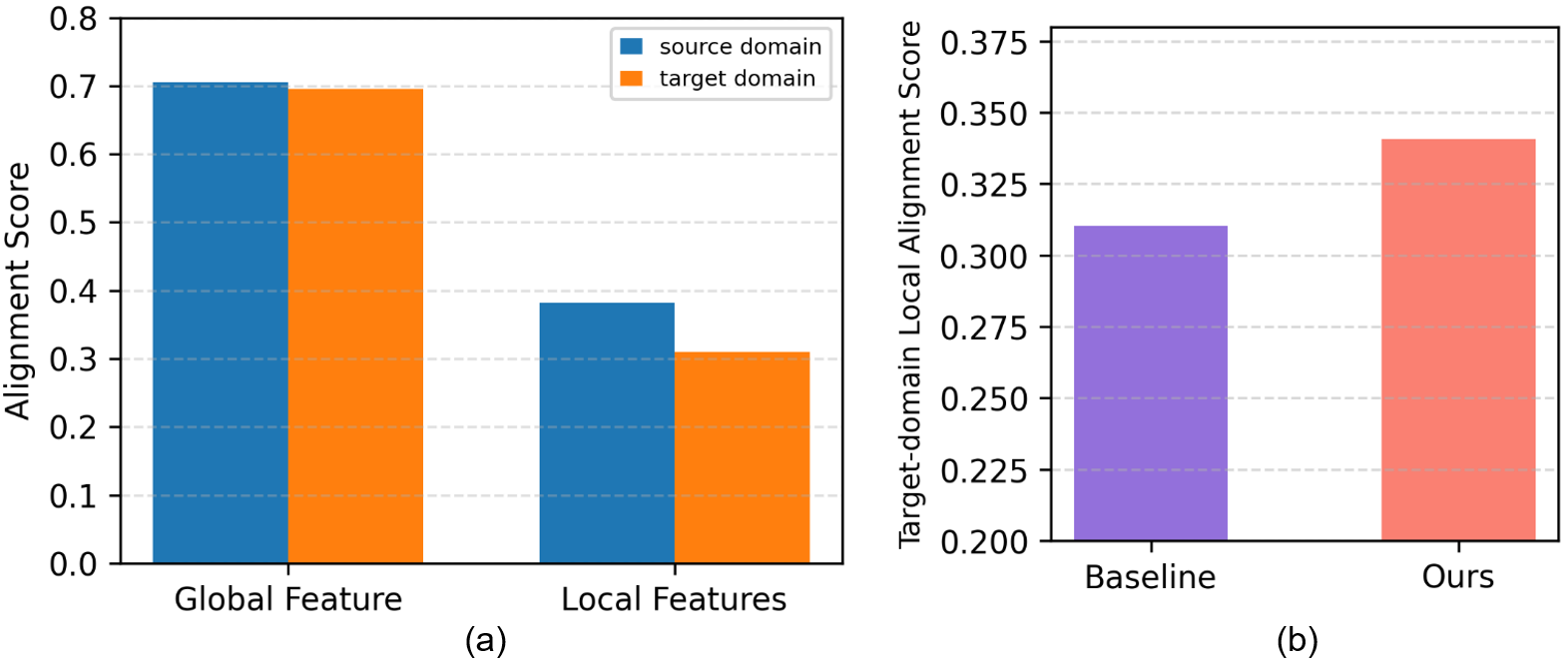} \vspace{-0.2cm}
\caption{(a) To validate our hypothesis, we measure the alignment score between global / local features and text features, and we find the alignments of both global and local features are harmed under the CDFSL task, but the local features show a larger decline in the alignment score, verifying our hypothesis. (b) Our proposed method effectively improves the local alignment in target domains. }
\label{alignment score}
\vspace{-0.5cm}
\end{figure}

Typical downstream domains, such as medical diagnosis, always rely on fine-grained visual cues for interpretable recognition.
For example, as illustrated in Fig.~\ref{attnmap}a, different categories of lung diseases in X-ray images may manifest as subtle abnormalities in texture or density within the lung area~\cite{wang2017chestx}, as shown by red boxes in Fig.~\ref{attnmap}a.
 However, in Fig.~\ref{attnmap}b, we find current CLIP models fine-tuned on target domains always fail to focus on these subtle visual patterns, where the heatmap can only coarsely outline the contour of the body  (e.g., the lung border or road edge), 
but cannot focus on these red boxes. Similar phenomena are also observed in other target-domain images. However, it may not be an issue in source-domain images (Fig.~\ref{attnmap}c), since the model can roughly highlight all important areas of the image.
Since some works have found CLIP's shortcomings in capturing subtle local patterns, we hypothesize that \textbf{the domain gap and scarce training data exacerbate CLIP's shortcomings in capturing subtle local patterns more than holistic patterns}.
To quantitatively validate this issue, we further measure the alignment scores~\cite{DBLP:journals/corr/abs-2406-17639} 
between CLIP's global features (CLS token), or local patch features, and their text semantics (Fig.~\ref{alignment score}a). 
The results show that the domain gap harms the alignment between visual and text features, where the local features are more vulnerable, verifying our hypothesis.

Therefore, in this paper, we aim to address the misalignment between local visual features and text features exacerbated by CDFSL (Fig.~\ref{alignment score}b). 
To achieve this goal, following the pretraining of CLIP, during the target-domain fine-tuning, we need to align each local visual feature (patch feature) with its corresponding text descriptions. However, 
firstly, image patches lack supervision that provides direct labels of texts,   
making it difficult to directly associate patch features with text semantics. Secondly, patches may contain noise or irrelevant regions (e.g., background) due to richer information in the visual modality, 
and there is insufficient supervision to remove the noise in few-shot scenarios.

To solve these problems, we turn to self-supervision information for the local feature alignments.
Inspired by translation \cite{DBLP:conf/nips/LiuBK17,DBLP:conf/eccv/LiHMLZJ18}, we propose a regularization framework based on cycle consistency, named CC-CDFSL, to explicitly constrain the bidirectional mapping between local visual features and text features, \textit{just like translating local visual features into text features and then translating them back into visual features (and vice versa)}. 
Specifically, we design two cycle paths: Text-to-Image-to-Text (T-I-T) and Image-to-Text-to-Image (I-T-I). 
To reduce the noise or irrelevant regions in the visual modality, we further propose a Semantic Anchor 
(SA)
mechanism, which first augments visual features to provide a larger corpus for the text-to-image mapping, and then shrinks the effective visual features to filter out irrelevant image-to-text mappings.
Additionally, these cycle paths also enhance interpretability by enabling visualization of patch-level semantic similarities, providing insights into learned patterns and model decisions for interpretable CDFSL.

In summary, our contributions are listed as follows:
\begin{itemize}
    \item To the best of our knowledge, we are the first to find and address the local-feature-alignment problem in the CLIP-based CDFSL scenarios.
    \item To address this problem, we propose the CC-CDFSL method to provide self-supervision between local visual features and text features, by translating local visual features into text features and then translating back into visual features (and vice versa), and constraining the original features close to the translated back features.
    \item To reduce the noise caused by the richer information in the visual modality, we further propose a Semantic Anchor 
    mechanism, which first augments the visual features and then shrinks them to improve the fine-grained mapping between modalities.
    \item Extensive experiments on various datasets, backbones, and fine-tuning methods
    demonstrate that CC-CDFSL consistently improves performance and outperforms state-of-the-arts with better patch-level interpretability.
\end{itemize}

%% file: sec/2_Related.tex
\section{Related Work}
\label{sec:formatting}
\textbf{Cross-Domain Few-Shot Learning}  
aims to enable a model trained on a data-rich source domain 
to rapidly adapt to a target domain.
The main challenge is the significant distribution difference 
between the source and target domains \cite{zou2024attention}.
Existing research primarily revolves around three perspectives: metric-based learning methods \cite{DBLP:conf/ijcai/XueHZH0SJ24} achieve efficient classification by constructing similarity metrics in a feature space; meta-learning methods \cite{DBLP:conf/iclr/TsengLH020, HuM22} optimize rapid adaptation by simulating few-shot tasks; and transfer learning methods \cite{PhooH21,LiangZDL21} leverage pre-trained models from the source domain, adapting to the target domain through fine-tuning or feature alignment. 
Now vision-language models (VLMs)  provide new opportunities for Source-Free Cross-Domain Few-Shot Learning~\cite{DBLP:journals/tip/XuLZFSCL24, 
zhang2026reclaiming} (SF-CDFSL). 
 
\vspace{0.1cm}
\noindent \textbf{Parameter-Efficient Fine-Tuning 
(PEFT) 
in VLMs}. 
Recent research on the efficient fine-tuning of vision-language large models focuses on three primary approaches. Prompt-based methods \cite{zhou2022coop, zhou2022cocoop, khattakMaPLe} freeze encoder parameters, introducing trainable vectors into text or visual inputs to guide task-specific outputs. Adapter-based methods \cite{CLIP-Adapter,Tip-Adapter} insert lightweight bottleneck modules into specific layers of pre-trained models, utilizing residual connections to integrate features for task adaptation. Low-Rank Adaptation (LoRA) methods \cite{hu2022lora,zanella2024low} employ low-rank matrix decomposition for weight updates, significantly reducing parameter counts while preserving inference efficiency. 
Although effective for in-domain fine-tuning~\cite{zhang2025decoupling}, these methods only align image-level CLS tokens with text, leaving local patches without explicit semantic supervision. Consequently, fine-grained cues are prone to being overlooked when transferring across domains (Figs.~\ref{attnmap} and  \ref{alignment score}).

\vspace{0.1cm}
\noindent \textbf{Local Alignment in VLMs}. 
Some approaches explore local alignment during the pretraining phase of VLMs,  such as transforming image embeddings  into region representations given spatial hints \cite{DBLP:journals/corr/abs-2410-02746}, introducing region-text  contrastive learning, or  hard fine-grained negative sample learning \cite{DBLP:journals/corr/abs-2505-05071}. 
 Additionally, recent advancements in downstream fine-tuning have increasingly focused on local alignment strategies to enhance fine-grained cross-modal understanding. Notable approaches include  employing region-based or local feature matching with LLM-generated  text descriptions to capture detailed semantics \cite{
 DBLP:conf/cvpr/ChoiJE25,  DBLP:conf/iclr/0002YSLR023} or refine the similarity score \cite{DBLP:conf/icml/0004LEF0L24}.
 However, these methods  often rely on explicit region annotations or additional auxiliary losses requiring extensive labeled data, which is impractical in CDFSL scenarios where annotated data is scarce. 
In contrast, local alignment in CDFSL remains unexplored, and we find that the domain gap and limited data worsen local visual feature misalignment more than global features. 
So we
 leverage self-supervised cycle consistency to achieve robust patch-level alignment,  which also enhances model interpretability.

%% file: sec/3_method.tex
\section{Method}
\subsection{Preliminaries}
\textbf{Cross-Domain Few-Shot Learning} 
(CDFSL) 
requires a model trained on a source domain (e.g., miniImageNet) to adapt to target domains (e.g., EuroSAT) using a few labeled samples (\textit{K-way N-shot} setting), and evaluates its classification performance on the target domain.
In \textit{K-way N-shot} classification tasks, the support set \(\mathcal{S}\) contains \(K\) classes with \(N\) labeled samples each (\(N\) is usually small)
, denoted as \(\mathcal{S} = \{(x_i, y_i)\}_{i=1}^{K \times N}\), where the label \(y_i \in \{1, \dots, K\}\). The query set \(\mathcal{Q} = \{(x_j^{q}, y_j^{q})\}\) assesses the model's  performance, and \(\mathcal{S} \cap \mathcal{Q} = \emptyset\). We mainly focus on the fine-tuning on the target domain with only $|\mathcal{S}|= N \cdot K$ samples.

\vspace{0.1cm}
\noindent \textbf{Visual-Language Models}
exemplified by CLIP \cite{radford2021learningtransferablevisualmodels}   have significantly advanced vision-related tasks.
CLIP uses a dual-encoder architecture with separate image and text encoders. The image encoder, which can be a ResNet \cite{he2016deep} or 
ViT
\cite{dosovitskiy2021an}, extracts features from an image \(I\) into a feature vector \(v_I = f_{\text{image}}(I) \in \mathbb{R}^d\), where \(d\) is the feature dimension. In the Vision Transformer, the input image \(I \in \mathbb{R}^{H \times W \times 3}\) is split into \(M = \left\lfloor \frac{H}{p} \right\rfloor \times \left\lfloor \frac{W}{p} \right\rfloor\) patches of size \(p \times p \times 3\). These patches are mapped to patch embeddings to form the initial image embedding sequence \(E_0 \in \mathbb{R}^{M \times d_v}\). After concatenating a class token \(c_0\) and adding position embeddings, the sequence is processed by \(L\) Transformer layers. The final class token \(c_L\) is projected to obtain \(v_I = P_v(c_L)\).

The text encoder processes the input text \(T'\) by converting it into a sequence of discrete tokens. Each token is embedded into a vector to form the initial text embedding \(T'_0 \in \mathbb{R}^{N \times d_t}\). Special beginning (\(b_0\)) and end (\(e_0\)) tokens are added, and position embeddings are included. The sequence is then processed by \(L\) Transformer layers. 
The final text feature vector \(v_T = f_{\text{text}}(T') = P_t(e_L)\) is obtained by projecting the output vector of the EOS token \(e_L\).

For  downstream image classification tasks with \(C\) categories and their corresponding text descriptions \(D_1, D_2, \dots, D_C\), the image \(I\) is classified into the category with the highest similarity. 
Under the standard CLIP fine-tuning framework, 
global feature \( G \in \mathbb{R}^d \)  generated by the CLS token is typically used for classification tasks. The original loss function is expressed as:
{
\vspace{-0.2cm}
\small
\begin{equation}
\mathcal{L}_{\text{CE}} = - \log p(y= c | x) = -\log \frac{\exp(\text{sim}(G, T_c) / \tau)}{\sum_{i=1}^C \exp(\text{sim}(G, T_i) / \tau)}
\label{ep:ce}
\end{equation}
}
\noindent Here, \( \tau \) is the temperature scaling parameter,
and \( T_c \) is the text feature corresponding to the true class label \( c \).

\begin{figure*}[t]
\centering
\includegraphics[width=0.98\textwidth]{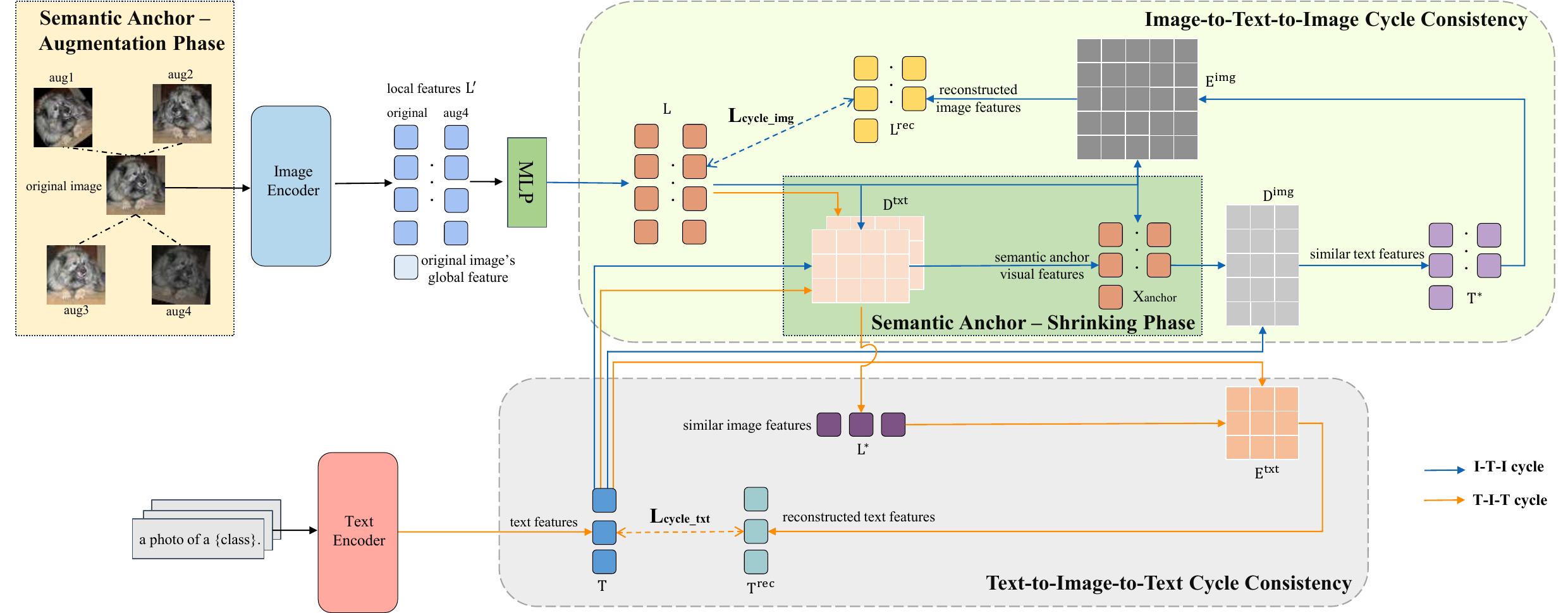}
\caption{Overview of our framework, consisting of three key components: (a) the Text-to-Image-to-Text (T-I-T) cycle-consistency module, (b) the Semantic Anchor (SA) module, and (c) the Image-to-Text-to-Image (I-T-I) cycle-consistency module. The process begins with the SA module augmenting raw images to create a larger corpus, followed by extracting local image features from these images and transforming them via MLP to align with the text feature space. The T-I-T cycle then uses text features to select semantically relevant patches and maps them back to reconstruct text features, enhancing local feature alignment. Subsequently, the SA module shrinks the feature set to select class-relevant anchor patches, which are used in the I-T-I cycle to map these anchor visual features through text features to augmented image features. Our model  improves local alignment and interpretability in cross-domain few-shot learning.}
\label{fig:cc}
\vspace{-0.3cm}
\end{figure*}

\subsection{CC-CDFSL}
In the ViT architecture used in the image encoder, the local features of image patches 
contain rich visual information such as edges, 
shapes, and colors \cite{zeng2025local},
 making them important for interpretable diagnosis in specialized domains like medical imaging.
As discussed earlier, CLIP pre-training relies solely on the CLS token to represent the whole image and leaves local patches unsupervised, leading to criticism of CLIP’s limited ability to capture fine-grained visual details \cite{
DBLP:journals/corr/abs-2505-05071}. 
We go one step further and find that downstream cross-domain tasks demand 
stronger local discrimination. For instance, Pneumothorax appears as a thin dark band along the lung border, while Nodules manifest as faint circular shadows in chest X-rays (Fig. \ref{attnmap}a). Fig. \ref{attnmap}b  shows that target-domain fine-tuning harms CLIP’s local features more severely, so we hypothesize that the domain gap and scarce training data exacerbate CLIP’s local feature misalignment more than global feature misalignment. To validate this, we quantitatively measure the alignment scores for global and local patch features on the support set: 
\vspace{-0.2cm}
{
\small
\begin{align}
\text{A}_{\text{g}} &= \frac{1}{|\mathcal{S}|}\sum_{i=1}^{|\mathcal{S}|}\frac{\mathbf{v}^{\text{cls}}_i\cdot\mathbf{t}_{y_i}}{\|\mathbf{v}^{\text{cls}}_i\|\|\mathbf{t}_{y_i}\|}, &
\text{A}_{\text{l}} &= \frac{1}{|\mathcal{S}| M}\sum_{i=1}^{|\mathcal{S}|}\sum_{j=1}^{M}
\frac{\mathbf{v}_{i,j}\cdot\mathbf{t}_{y_i}}{\|\mathbf{v}_{i,j}\|\|\mathbf{t}_{y_i}\|}
\label{eq:align_global_local}
\vspace{-0.3cm}
\end{align}
}
\noindent where 
$M$ is the number of patches per image,
and $\mathbf{t}_{y_i}\in\mathbb{R}^{d}$ is the normalized text feature for the true class label $y_i$. 

Fig.~\ref{alignment score}a shows
that in cross-domain tasks, the decrease in $\text{A}_{\text{l}}$ is significantly greater than that of $\text{A}_{\text{g}}$, confirming that domain discrepancies amplify CLIP’s inability to capture fine-grained discriminative information.

Drawing inspiration from this limitation, this study introduces a novel self-supervision strategy based on cycle consistency. By explicitly constraining the bidirectional mapping relationship between local features and text semantics, we effectively improve the local alignment between visual and textual modalities, and enhance interpretability. 
The overall framework of our method is shown in Figure \ref{fig:cc}.

\vspace{0.1cm}
\noindent\textbf{Semantic Anchor Module: Augmentation Phase}.
To reduce noise and irrelevant regions in the visual modality, we introduce the Semantic Anchor (SA) mechanism, operating in an \emph{augment-then-shrink} manner. In the augmentation phase, we generate $A$ 
augmented views for each image and include the original image, flattening all patch features into a single augmented feature matrix:
\begin{equation}
\mathbf{X}_{\text{aug}}\in\mathbb{R}^{((A + 1)\cdot M)\times d}
\end{equation}

This enlarges the candidate corpus for the first hop of the T-I-T cycle. 
Next, we concatenate these features across the entire support set to form the local feature matrix:
\vspace{-0.1cm}
{
\small
\setlength{\arraycolsep}{1pt} 
\begin{equation}
\mathbf{L}' = \text{Concat} \left( \{\mathbf{X}_{\text{aug}}^{(i)} \mid i \in \mathcal{S} \} \right) \in \mathbb{R}^{H \times d}, \quad H = |\mathcal{S}|(A+1)M
\end{equation}
}  

\vspace{0.1cm}
\noindent\textbf{Local Features Transformation}.
Since CLIP pre-training uses global features to align visual and textual modalities, directly using local image features may result in a significant modality gap. Thus, we first transform local image features 
$\mathbf{L}' \in \mathbb{R}^{H \times d}$
generated from all patches across the support set to reduce the distribution discrepancy with the textual feature space via a two-layer MLP:
\begin{equation}
\mathbf{L} = \mathrm{MLP}(\mathbf{L}') = \mathrm{ReLU}(\mathbf{L}'\cdot W_1) \cdot W_2 
\label{eq32}
\end{equation}

\vspace{0.1cm}
\noindent\textbf{Text-to-Image-to-Text Cycle Consistency}.
In the proposed  T-I-T cycle, we first compute the cosine  similarity matrix $\mathbf{D}^{\text{txt}} \in \mathbb{R}^{C \times H}$ between  text features $ \mathbf{T} \in \mathbb{R}^{C \times d}$  (\( C \) is the number of categories) and local image features:
\vspace{-0.2cm}

{
\small
\begin{equation}
\mathbf{D}^{\text{txt}}_{j,i} = \frac{\mathbf{T}_j \cdot \mathbf{L}_i}{\|\mathbf{T}_j\| \|\mathbf{L}_i\|}, \quad \mathbf{D}^{\text{txt}}_{[b]} \in \mathbb{R}^{C \times M}, \quad b \in [1,2,... , |\mathcal{S}|(A+1)]
\end{equation}
}

\begin{table*}[ht]
\centering
\small          
\setlength{\abovecaptionskip}{2pt}  
\caption{Accuracies (\%) of target domain datasets of 5-way 1-shot and 5-shot tasks. Refer to the Appendix for the extended table.
}
\label{tab:main_results}   
\begin{adjustbox}{max width=0.9\textwidth}
\begin{threeparttable}
\arrayrulecolor{black}\arrayrulewidth=0.6pt
\begin{tabular}{l|l|c|c|cccc|c|}  
\hline
\textbf{Task} & \textbf{Method} & \textbf{Mark} & \textbf{backbone} & \textbf{ISIC} & \textbf{ChestX} & \textbf{EuroSAT} & \textbf{CropDisease} & \textbf{Avg} \\ \hline  

\multirow{15}{*}{\rotatebox{90}{\textbf{5-way 1-shot}}}  
& StepSPT~\cite{DBLP:journals/corr/abs-2411-10070} &  TPAMI-25 & ViT/CLIP & 32.97 & \textbf{22.84} & 70.01 & 84.84 & 52.68 \\ 
& Tip-Adapter~\cite{Tip-Adapter} & ECCV-22 & ViT/CLIP & 32.68 & 22.24 & 75.44 & 77.15 & 51.87 \\
& AMU-Tuning~\cite{DBLP:conf/cvpr/TangLW0H24} &  CVPR-24 & ViT/CLIP & 32.29 & 21.56 & 72.24 & 80.20 & 51.57 \\
& LP++~\cite{lp2024} & CVPR-24 & ViT/CLIP & 33.63 & 21.72 & 73.05 & 81.84 & 52.56 \\
& LDC~\cite{Li_2025_CVPR} & CVPR-25 & ViT/CLIP & 33.72 & 22.32 & 74.39 & 84.07 & 53.62 \\
& CoOp~\cite{zhou2022coop} & IJCV-22 & ViT/CLIP & 29.47 & 20.95 & 68.16 & 79.27 & 49.46 \\
& \cellcolor{cyan!10}\textbf{CoOp + OURS}  & \cellcolor{cyan!10} - & \cellcolor{cyan!10}ViT/CLIP & \cellcolor{cyan!10}31.96 & \cellcolor{cyan!10}21.37 & \cellcolor{cyan!10}71.17 & \cellcolor{cyan!10}81.80 & \cellcolor{cyan!10}51.58 \\
& CLIP-Adapter~\cite{CLIP-Adapter} & IJCV-24 & ViT/CLIP & 30.52 & 21.32  & 67.87 & 82.05 & 50.44 \\
&
\cellcolor{cyan!10}\textbf{CLIP-Adapter + OURS} & \cellcolor{cyan!10}- & \cellcolor{cyan!10}ViT/CLIP & \cellcolor{cyan!10}32.47 & \cellcolor{cyan!10}22.14 & \cellcolor{cyan!10}72.89 & \cellcolor{cyan!10}82.01 & \cellcolor{cyan!10}52.37\\
& Maple~\cite{khattakMaPLe}  & CVPR-23 & ViT/CLIP & 33.56 & 20.89 & 75.05 & 82.67 & 53.04 \\
& \cellcolor{cyan!10}\textbf{Maple + OURS} & \cellcolor{cyan!10}- & \cellcolor{cyan!10}ViT/CLIP & \cellcolor{cyan!10}34.00 & \cellcolor{cyan!10}21.41 & \cellcolor{cyan!10}78.92 & \cellcolor{cyan!10}85.21 & \cellcolor{cyan!10}54.89\\
& CLIP-LoRA~\cite{zanella2024low} & CVPR-24 &  RN50/CLIP & 32.01 & 21.76 & 57.79& 65.24 & 44.20  \\    
& \cellcolor{cyan!10}\textbf{CLIP-LoRA + OURS} & \cellcolor{cyan!10}- & \cellcolor{cyan!10} RN50/CLIP & \cellcolor{cyan!10}35.21 & \cellcolor{cyan!10}\underline{22.75} & \cellcolor{cyan!10}59.23 & \cellcolor{cyan!10}72.85 & \cellcolor{cyan!10}47.51 \\   
\cline{2-9}  
& CLIP-LoRA~\cite{zanella2024low} & CVPR-24 & ViT/CLIP & 35.23 & 21.73 & 81.49 & 85.11 & 55.89 \\
& \cellcolor{cyan!10}\textbf{CLIP-LoRA + OURS} & \cellcolor{cyan!10}- & \cellcolor{cyan!10}ViT/CLIP & \cellcolor{cyan!10}\textbf{38.13} & \cellcolor{cyan!10}22.21 & \cellcolor{cyan!10}\textbf{86.07} & \cellcolor{cyan!10}\textbf{88.91} & \cellcolor{cyan!10}\textbf{58.83} \\
& \hspace{1.8em}\textcolor{teal}{\small		\textbf{$\Delta$}} 	& - & - &	\textcolor{teal}{\small	\textbf{+2.90}} & 	\textcolor{teal}{\small	\textbf{+0.48}} & 		\textcolor{teal}{\small		\textbf{+4.58}} & 		\textcolor{teal}{\small		\textbf{+3.80}} & 		\textcolor{teal}{\small		\textbf{+2.94}} \\

\hline
\hline
\multirow{15}{*}{\rotatebox{90}{\textbf{5-way 5-shot}}}  
& StepSPT~\cite{DBLP:journals/corr/abs-2411-10070} &  TPAMI-25 & ViT/CLIP & 52.12 & \textbf{26.36} & 89.40 & 96.01 & 65.97 \\
& Tip-Adapter~\cite{Tip-Adapter} &  ECCV-22 & ViT/CLIP & 46.96 & 24.07 & 87.24 & 94.19 & 63.12 \\
& AMU-Tuning~\cite{DBLP:conf/cvpr/TangLW0H24} & CVPR-24 & ViT/CLIP & 44.60 & 23.34 & 88.47 & 94.26 & 62.66 \\
& LP++~\cite{lp2024} & CVPR-24 & ViT/CLIP & 48.49 & 23.89 & 87.48 & 94.47 & 63.58 \\
& LDC~\cite{Li_2025_CVPR} &  CVPR-25 & ViT/CLIP & 49.70 & 25.89 & 90.82 & 96.71 & 65.78 \\
& CoOp~\cite{zhou2022coop}  & IJCV-22 & ViT/CLIP & 42.56 & 22.22 & 85.53 & 93.48 & 60.95 \\
& \cellcolor{cyan!10}\textbf{CoOp + OURS} & \cellcolor{cyan!10}- & \cellcolor{cyan!10}ViT/CLIP & \cellcolor{cyan!10}42.75 & \cellcolor{cyan!10}22.76 & \cellcolor{cyan!10}86.43 & \cellcolor{cyan!10}93.82 & \cellcolor{cyan!10}61.44 \\
& CLIP-Adapter~\cite{CLIP-Adapter} & IJCV-24 & ViT/CLIP & 44.09 & 23.53 & 84.48 & 93.34 & 61.36 \\
&
\cellcolor{cyan!10}\textbf{CLIP-Adapter + OURS} & \cellcolor{cyan!10}- & \cellcolor{cyan!10}ViT/CLIP & \cellcolor{cyan!10}46.67 & \cellcolor{cyan!10}23.87  & \cellcolor{cyan!10}86.99 & \cellcolor{cyan!10}93.89 & \cellcolor{cyan!10}62.86
\\
& Maple~\cite{khattakMaPLe} & CVPR-23 & ViT/CLIP & 46.72 & 22.29 & 89.29 & 93.51 & 62.95  \\
& \cellcolor{cyan!10}\textbf{Maple + OURS} & \cellcolor{cyan!10} - & \cellcolor{cyan!10}ViT/CLIP & \cellcolor{cyan!10}48.77 & \cellcolor{cyan!10}23.05 & \cellcolor{cyan!10}91.72 & \cellcolor{cyan!10}95.16 & \cellcolor{cyan!10}64.67 
\\
& CLIP-LoRA~\cite{zanella2024low} \phantom{zhengzaipao} & CVPR-24 & RN50/CLIP & 46.53 & 23.00 & 77.26 & 87.42 & 58.55 \\ 
& \cellcolor{cyan!10}\textbf{CLIP-LoRA + OURS} & \cellcolor{cyan!10}- & \cellcolor{cyan!10} RN50/CLIP & \cellcolor{cyan!10}49.11 & \cellcolor{cyan!10}23.55 & \cellcolor{cyan!10} 79.23& \cellcolor{cyan!10}90.99 & \cellcolor{cyan!10}60.72   \\ 
\cline{2-9}  
& CLIP-LoRA~\cite{zanella2024low} & CVPR-24 & ViT/CLIP & 50.68 & 24.44 & 92.63 & 96.20 & 65.99 \\ 
& \cellcolor{cyan!10}\textbf{CLIP-LoRA + OURS} & \cellcolor{cyan!10}- & \cellcolor{cyan!10}ViT/CLIP & \cellcolor{cyan!10}\textbf{54.72} & \cellcolor{cyan!10}25.47 & \cellcolor{cyan!10}\textbf{94.35} & \cellcolor{cyan!10}\textbf{97.08} & \cellcolor{cyan!10}\textbf{67.90} \\ 
& \hspace{1.8em}\textcolor{teal}{\small\textbf{$\Delta$}}  & - & - &
\textcolor{teal}{\small\textbf{+4.04}} & \textcolor{teal}{\small\textbf{+1.03}} &
\textcolor{teal}{\small\textbf{+1.72}} & \textcolor{teal}{\small\textbf{+0.88}} &
\textcolor{teal}{\small\textbf{+1.91}} \\
\hline
\end{tabular}
\end{threeparttable}
\end{adjustbox}
\vspace{-0.4cm}
\end{table*}

All features are processed to be L2-normalized for brevity in the rest, so cosine similarity is expressed as the dot product.
Next, for each text feature $\mathbf{T}_j$, we select the most relevant local image feature to construct the similar image feature matrix $\mathbf{L}^* \in \mathbb{R}^{C \times d}$:
\begin{equation}
\mathbf{L}^*_j = \mathbf{L}_{\operatorname{arg}\max_{i} \mathbf{D}^{\text{txt}}_{j,i}} \in \mathbb{R}^d
\end{equation}

Then, we compute the reverse similarity matrix $\mathbf{E}^{\text{txt}}$:
\begin{equation}
\mathbf{E}^{\text{txt}} = \mathbf{L}^* \cdot \mathbf{T}^\top  \in \mathbb{R}^{C \times C}
\end{equation}
\noindent where 
$\mathbf{E}^{\text{txt}}_{j,k}$ represents the cosine similarity 
between the image patch feature $\mathbf{L}^*_j$ (selected as the most relevant to the $j$-th text feature $\mathbf{T}_j$) and the $k$-th text feature $\mathbf{T}_k$.

The cycle consistency loss is constructed by maximizing the cosine similarity between each text feature $\mathbf{T}_j$ and the reconstructed text feature $\mathbf{T}^\text{rec}_j \in \mathbb{R}^{d}$:
{
\vspace{-0.2cm}
\begin{equation}
\mathcal{L}_{\text{cyc\_txt}} = 1 - \frac{1}{C}\sum_{j=1}^{C}
\text{sim}(\mathbf{T}_j,\, \mathbf{T}^\text{rec}_j)
\end{equation}
}
where $\mathbf T^\text{rec}_j = \mathbf T_{\,\operatorname{arg}\max_{k} \mathbf E^{\text{txt}}_{j,k}} $ is 
the text feature
corresponding to 
the $j$-th text feature after reverse mapping. 
 This process encourages 
the model to learn more discriminative and semantically meaningful local 
representations (Fig.~\ref{t2t}). 

 
\vspace{0.1cm}
\noindent\textbf{Semantic Anchor Module: Shrinking Phase}.
Before the I-T-I cycle, we filter out
noisy or irrelevant patches 
to reduce semantic drift 
by selecting the most class-relevant anchor features. 
For the \(b\)-th image and
the \(j\)-th class, 
we pick the top-$k$ indices of the most similar patches:
\vspace{-0.2cm}
\begin{equation}
\mathcal{I}^{(b)}_j = \operatorname{top-k}_i \bigl( \mathbf{D}^{\text{txt}}_{[b], j, i} \bigr)
\label{eq:k}
\end{equation}

We merge and deduplicate to obtain the anchor  index set:
{
\small
\begin{equation}
\mathcal{I}_{\text{anchor}} = \operatorname{unique}\left( \bigcup_{b=1}^{|\mathcal{S}|(A+1)} \bigcup_{j=1}^{C} \mathcal{I}^{(b)}_j \right), \quad V = \left| \mathcal{I}_{\text{anchor}} \right|
\label{eq:anchor}
\end{equation}
}

The anchor visual features are extracted as:
\begin{equation}
\mathbf{X}_{\text{anchor}} = \mathbf{L}_{\mathcal{I}_{\text{anchor}}} \in \mathbb{R}^{V \times d}
\end{equation}

These anchor features represent the local features most relevant to class semantics, suppressing 
low-information regions (Fig.~\ref{select}), and are used exclusively in the I-T-I cycle. 

\vspace{0.1cm}
\noindent\textbf{Image-to-Text-to-Image Cycle Consistency}.
Mirroring  the above T-I-T consistency process but in another direction,  we use text features as an intermediary to achieve cyclic mapping from original to augmented image features.

Specifically, for each anchor $\mathbf{x}_n \in \mathbf{X}_{\text{anchor}}$, we leverage text features to bridge original and its augmented image features:

    1. Identify the most similar text  feature:
    \begin{equation}
    t_n = \mathbf{T}_{\operatorname{argmax}_j (\mathbf{x}_n \cdot \mathbf{T}_j)} \in \mathbb{R}^{d}
    \end{equation}
    
    2. Using the selected text feature $t_n$ as an intermediary, retrieve the most similar patch only in the augmented feature space of $\mathbf x_n$ (i.e., $\mathbf{X}_{\text{aug}}$ ) to complete the cyclic mapping:
\begin{equation}
m^* = \operatorname{argmax}_m (t_n \cdot \mathbf{X}_{\text{aug},m}), \quad \hat{\mathbf{x}}_n = \mathbf{X}_{\text{aug}, m^*}\in \mathbb{R}^{d}
\end{equation}

We compute the cosine distance between anchors and their retrieved counterparts to construct the image-to-text-to-image cycle consistency loss:
\begin{equation}
\mathcal{L}_{\text{cyc\_img}} = 1 - \frac{1}{V} \sum_{n=1}^{V} \text{sim}(\mathbf{x}_n, \hat{\mathbf{x}}_n)
\end{equation}

 This approach ensures reliable localization of local image features in the augmented space, enhancing the model's robustness to input variations (Tab.~\ref{tab:ret}).
 
\vspace{0.1cm}
\noindent\textbf{Consistency Loss}.
Finally, the total loss combines the original cross-entropy loss
  with both the  text-to-text  and  image-to-image cycle consistency losses:
\begin{equation}
\mathcal{L}_{\text{total}} = \mathcal{L}_{\text{CE}} + \lambda_1 \mathcal{L}_{\text{cyc\_txt}} + \lambda_2 \mathcal{L}_{\text{cyc\_img}}
\label{eq:loss}
\end{equation}
 where $\lambda_1$ and $\lambda_2$ are balancing hyperparameters.
 During inference, we assign the label to each query sample by computing the highest cosine similarity between its visual feature and the text features of all classes. 

%% file: sec/4_exp.tex
\section{Experiments}
\subsection{Implementation Details}
According to the benchmark provided by \cite{guo2020broader}, our model is  directly fine-tuned on four target domain datasets: CropDiseases \cite{mohanty2016using} (plant diseases), EuroSAT \cite{helber2019eurosat} (satellite imagery), ISIC2018 \cite{codella2019skin} (skin lesions), and ChestX \cite{wang2017chestx} (chest X-rays).
Following \cite{DBLP:journals/corr/abs-2411-10070}, all experiments in the main paper are conducted based on the CLIP model 
with the ViT-Base/16 backbone. Performance evaluations with other backbone variants are detailed in Supplementary Table~\ref{tab:vit}.
For all baseline methods, we adhere to the experimental settings reported in their respective original publications. 
For a fair comparison, all models are fine-tuned for 100 epochs 
on a single NVIDIA RTX 4090 GPU. 
Evaluation is conducted 100 times under the 1-shot setting and 400 times under the 5-shot setting, with the mean classification accuracy reported.
For the hyper-parameters in Eq.~\ref{eq:loss}, we set  $\lambda_1$  and  $\lambda_2$  based on empirical observation on a validation set via grid search; see the Appendix for details. 
$k$  in the Semantic Anchor module is fixed to 10 across all experiments. 
\subsection{Comparison with State-of-the-Art Methods}
We evaluate our  CC-CDFSL framework in comparison with the most competitive state-of-the-art (SOTA) methods.
These competing approaches cover representative technical paradigms in PEFT of vision-language models, including prompt learning (CoOp~\cite{zhou2022coop}) in  the text
 branch, adapter-based methods (Tip-Adapter~\cite{Tip-Adapter}, CLIP-Adapter~\cite{CLIP-Adapter}), multi-modal prompt learning (Maple~\cite{khattakMaPLe}), and low-rank adaptation (CLIP-LoRA~\cite{zanella2024low})
 —as well as recent advanced tuning strategies (AMU-Tuning~\cite{DBLP:conf/cvpr/TangLW0H24}, LP++~\cite{lp2024}, LDC~\cite{Li_2025_CVPR}, StepSPT~\cite{DBLP:journals/corr/abs-2411-10070}). 
As presented in Tab.~\ref{tab:main_results}, our CC-CDFSL consistently enhances the performance of all baseline methods.
\begin{figure}[t]
\centering
\includegraphics[width=0.76\columnwidth]{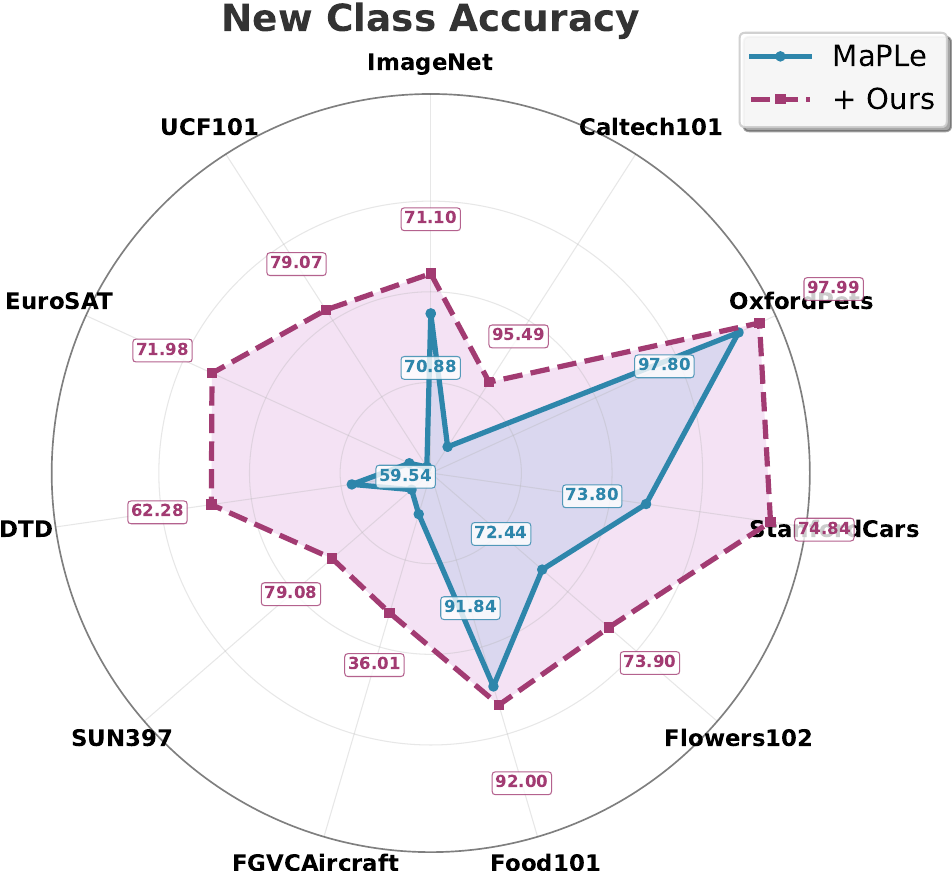}  
\caption{
Base-to-new generalization on 11 datasets.}
\label{Base-to-new generalization on 11 datasets}
\vspace{-0.4cm}
\end{figure} 
To investigate whether the proposed CC-CDFSL can be re-purposed beyond cross-domain few-shot learning, we further evaluate it on the widely used base-to-new generalization setting \cite{zhou2022cocoop, DBLP:conf/cvpr/GuoG25, khattakMaPLe} across 11  commonly  adopted datasets. As 
 illustrated 
in Figure~\ref{Base-to-new generalization on 11 datasets}, we achieve substantial gains on challenging fine-grained datasets, such as EuroSAT (+3.6\% in New), StanfordCars (+1.0\% in New), and DTD (+2.7\% in New).
These results indicate that cycle-consistent patch-level alignment not only mitigates domain shift in few-shot settings but also enhances generalization to unseen classes by learning semantically richer local representations. 
\subsection{Ablation Study}
\textbf{The effects of each component}. 
According to Table~\ref{abla},  both cycles 
boost  performance over the baseline, but the 
T-I-T cycle outperforms the I-T-I cycle.
We identify two primary reasons for this: First, using the StanfordCars dataset as an example, where category names are vehicle names, image patches (e.g., tires or headlights) struggle to accurately match corresponding text features in few-shot scenarios due to the limited number of categories,
resulting in text features that fail to effectively guide visual feature learning. 
Second,  the noise or redundant information
in image local features is difficult to completely eliminate, 
causing I-T-I reconstructed image features to deviate from their original semantics, whereas T-I-T focuses on the most semantically relevant patches, reducing interference from irrelevant patches and enhancing the robustness of local 
alignment.


\vspace{0.1cm}
\noindent\textbf{Retrieval strategies in I-T-I cycle}. Tab.~\ref{tab:ret} presents the impact of different retrieval strategies on model performance. 
Cross-view retrieval, by using text features as an intermediary to match semantic anchor features only in the augmented image feature space, significantly enhances the model’s robustness to input transformations (e.g., rotation, flipping), adapting to distribution differences in cross-domain few-shot learning. In contrast, intra-image retrieval  and all-images  retrieval are limited to the original image feature space and overly large selection range separately, lacking the diversity of augmented views and being susceptible to noise, resulting in less accurate semantic alignment.

\begin{table}[t]
\setlength{\abovecaptionskip}{2pt}  
\caption{Ablation study of the 5-way 5-shot task.}
\centering
\small
\setlength{\tabcolsep}{1mm}
\resizebox{0.4\textwidth}{!}{
\begin{tabular}{ccc*{5}{c}}
\toprule
T-I-T & I-T-I & SA & ISIC & ChestX & EuroSAT & Crop. & Ave. \\
\midrule
 &  &  & 50.68 & 24.44 & 92.63 & 96.20 & 65.98 \\
\checkmark &  &  & 51.13 & 25.15 & 93.79 & 96.37 & 66.61 \\
\checkmark &  & \checkmark & 54.30 & 25.35 & 94.33 & 96.95 & 67.73 \\
 & \checkmark & \checkmark & 53.81 & 25.14 & 93.83 & 97.01 & 67.45 \\
\checkmark & \checkmark & \checkmark &  \textbf{54.72} &  \textbf{25.47} &  \textbf{94.35} &  \textbf{97.08} &  \textbf{67.90} \\
\bottomrule
\end{tabular}
}
\label{abla}
\vspace{-0.25cm}
\end{table}

\begin{table}[t]
\setlength{\abovecaptionskip}{2pt}  
\caption{Impact of different retrieval strategies.}
\centering
\small
\setlength{\tabcolsep}{1.1mm}
\begin{tabular}{lccccc}
\toprule
Retrieval & ISIC & ChestX & EuroSAT & Crop. & Ave. \\
\midrule
Cross-view  & 53.77 & 25.13 & 93.75 & 96.84 & 67.37\\
Intra-image & 53.59 & 25.01 & 93.63 & 96.77 & 67.25 \\
All-images  & 53.66 & 25.02 & 93.46 & 96.64 & 67.19 \\
\bottomrule
\end{tabular}
\label{tab:ret}
\vspace{-0.5cm}
\end{table}

\begin{figure}[htbp]
\centering
\includegraphics[width=0.95\columnwidth]{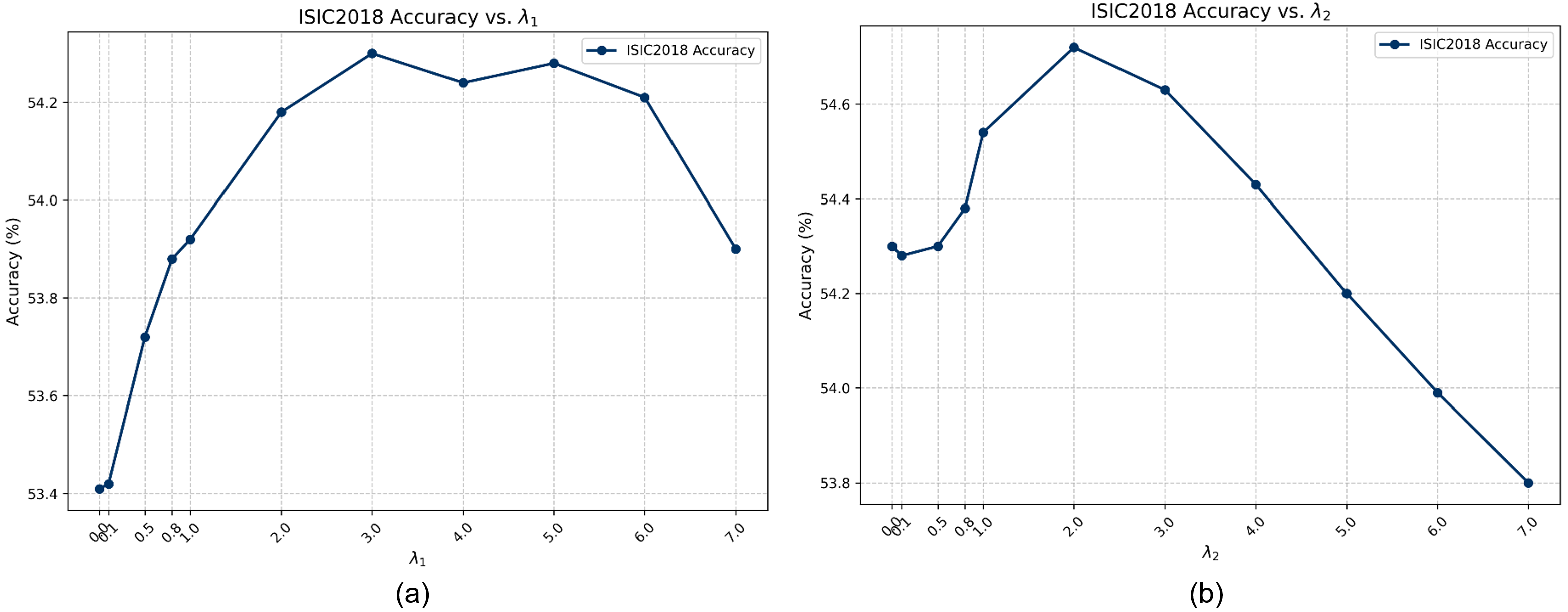} \vspace{-0.2cm}
 \caption{ Ablation study on hybrid  coefficient 
 in Eq.~\ref{eq:loss}. }
\label{fig:two_images}
\vspace{-0.2cm}
\end{figure}

\vspace{0.1cm}
\noindent\textbf{Hybrid coefficient $\lambda_1$ and $\lambda_2$}
balance the contributions of T-I-T and I-T-I cycle 
losses, respectively. We conducted a grid search over $[0.0, 7.0]$, finding that $\lambda_1 = 3.0$ and $\lambda_2 = 2.0$ yield the best 
accuracy  on  ISIC2018
(see Fig.~\ref{fig:two_images}). As shown in the line graph, increasing $\lambda_1$ enhances the T-I-T path's contribution, significantly improving performance.
However, when $\lambda_1 > 3.0$, performance plateaus, indicating that excessive focus on local alignment may neglect global features. The effect of $\lambda_2$ is similar but more
pronounced,  
 as the I-T-I path is more 
susceptible 
to noise.
\vspace{-0.1cm}

\begin{figure}[t]
\centering
\includegraphics[width=0.75\columnwidth]{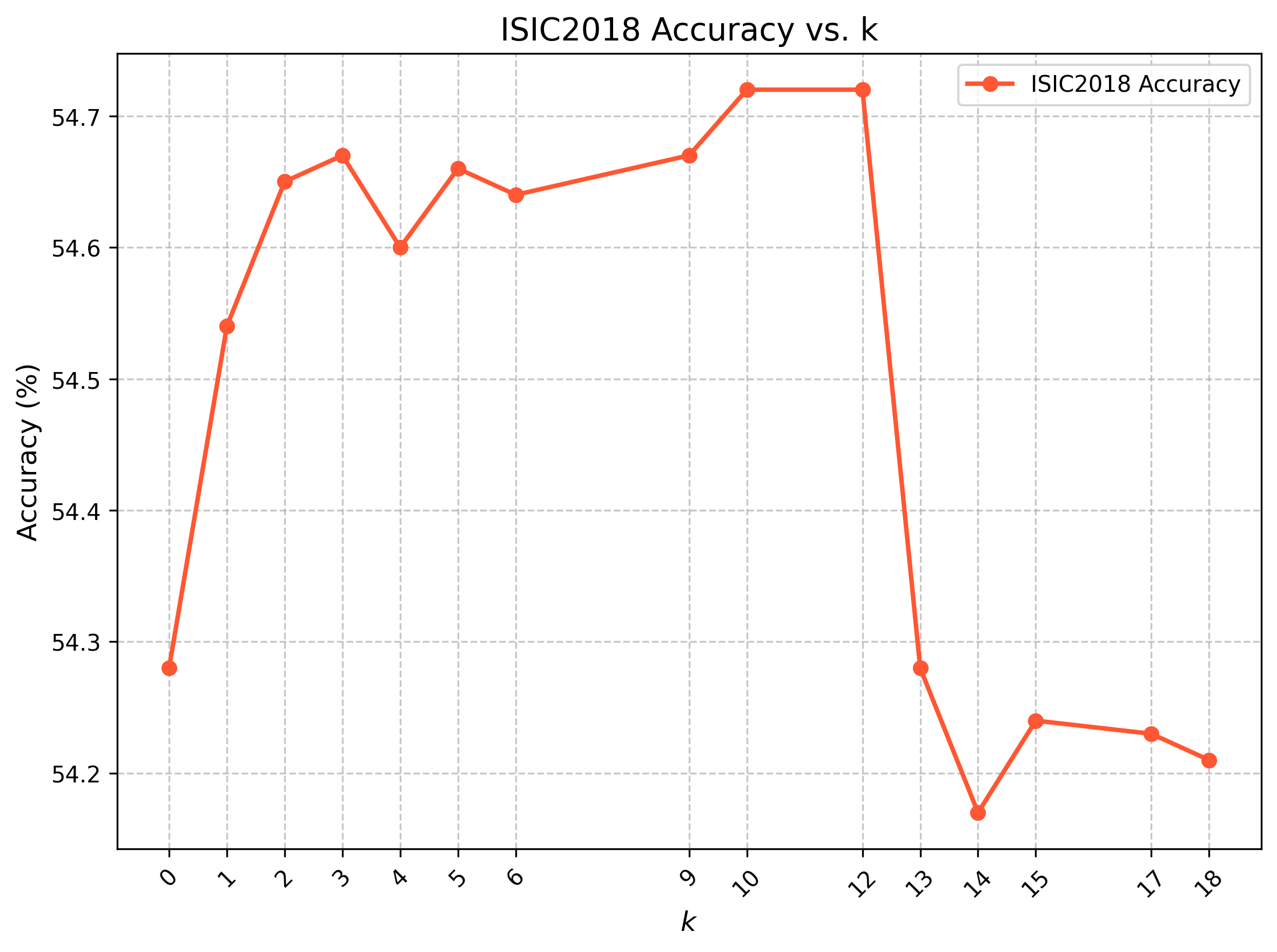} \vspace{-0.4cm}
 \caption{ Ablation study on hyperparameter $k$ in Eq.~\ref{eq:k}.  }
\label{fig:two_imagess}
\vspace{-0.4cm}
\end{figure}

\vspace{0.1cm}
\noindent\textbf{Anchor Patch Number \textit{k} in Eq.~\ref{eq:k}}.
In Fig.~\ref{fig:two_imagess}, we
investigate the influence of the anchor-patch budget $k$ (as defined in Eq.~\ref{eq:k}, controlling the number of patches that survive the Semantic Anchor shrinking phase and enter the I-T-I cycle for each image-class pair) on 5-way 5-shot accuracy. Results demonstrate that a small  $k (\leq3)$ under-utilizes informative cues, reducing the average accuracy from 54.67\% to 54.28\%, whereas an excessively large $ k (\geq13)$ re-introduces noisy regions and pushes accuracy back down to 54.21\%. Setting $k=10$ strikes an optimal balance, preserving fine-grained semantics while suppressing distractors and yielding the peak 54.72\% on the ISIC2018 dataset. Therefore, we adopt $k=10$ throughout all main experiments across the four datasets. 

\vspace{-0.1cm}
\subsection{Meaningful Patch  Selection in  SA}
\begin{figure}[htbp]
\centering
\includegraphics[width=0.88\columnwidth]{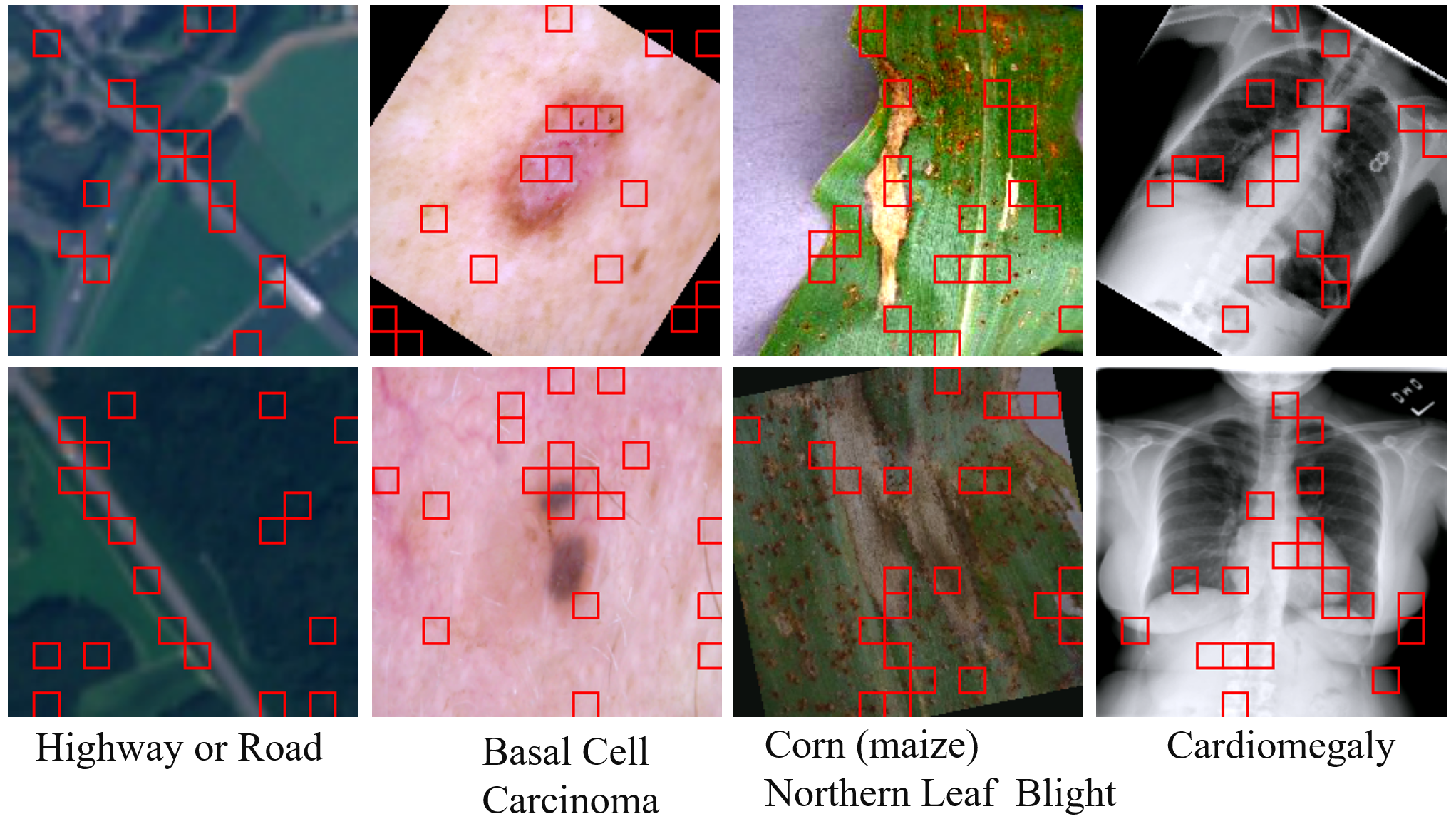} \vspace{-0.2cm}
\caption{
Semantic Anchor (SA) selection across four domains.
Red boxes indicate the top-$k\!=\!10$ patch proposals retained by the SA shrinking phase (Eq.~\ref{eq:anchor}).}
\label{select}
\vspace{-0.4cm}
\end{figure}

\vspace{-0.1cm}
As shown in Fig.~\ref{select}, the 
Semantic Anchor 
module only retains the top-k most reliable patch-text matches, filtering out ambiguous mappings and reducing semantic drift. 

\vspace{-0.1cm}
\subsection{Improved Local Alignment}
\begin{figure}[htbp]
\centering
\includegraphics[width=0.96\columnwidth]{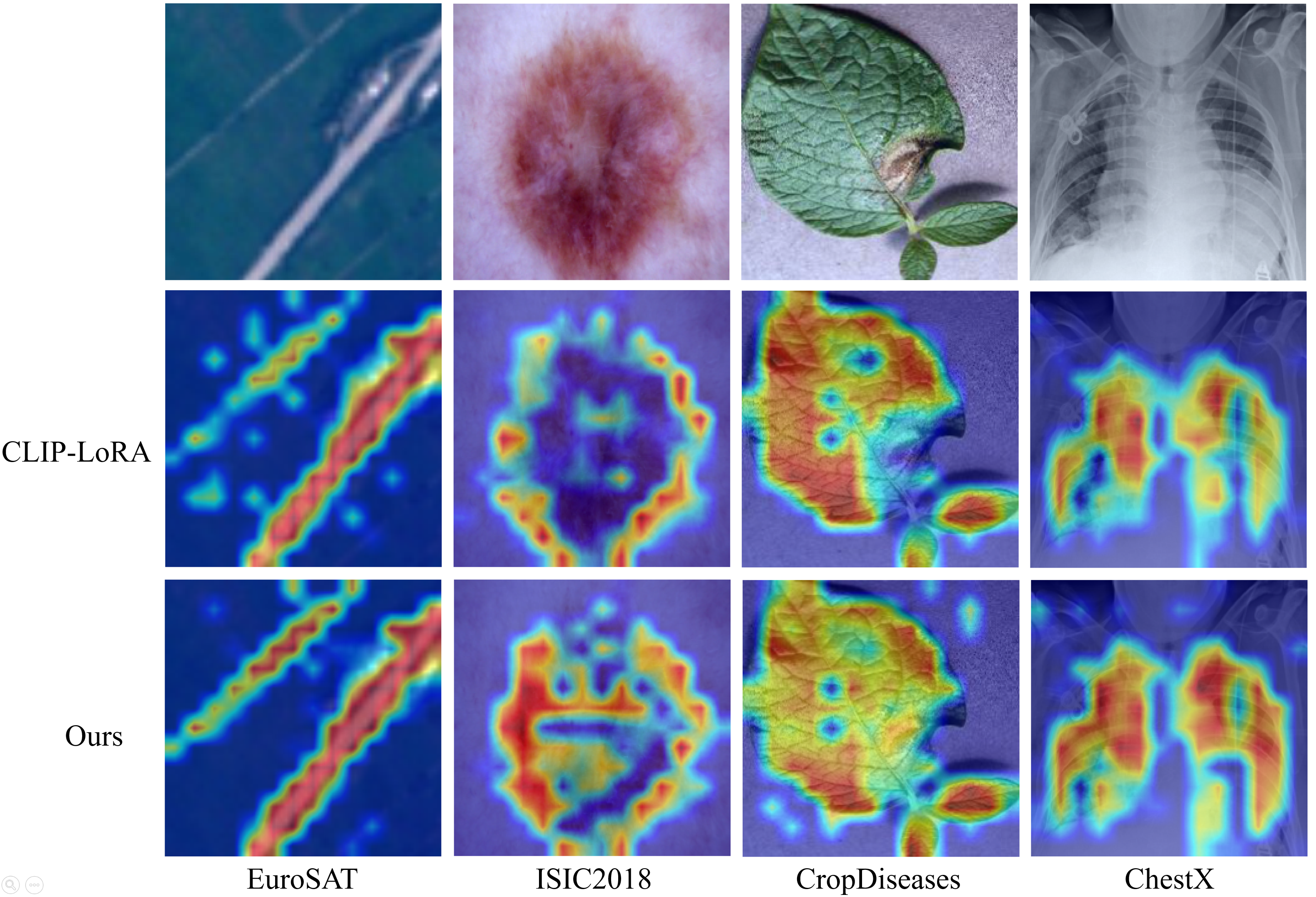} \vspace{-0.2cm}
\caption{
Our method (the third row) outperforms the baseline (the second row) in highlighting key details, offering a clearer view of important features in data-scarce scenarios.}
\label{attn}
\vspace{-0.4cm}
\end{figure}

\vspace{-0.1cm}
\noindent
Fig.~\ref{attn} 
illustrates 
the heatmaps of the baseline model and our 
CC-CDFSL 
across four datasets using Grad-CAM \cite{DBLP:conf/iccv/SelvarajuCDVPB17}. The comparison reveals that our method 
reduces interference from irrelevant features, highlighting image regions most relevant to the text.
This demonstrates that leveraging textual semantic guidance significantly enhances the model's capacity to concentrate on discriminative features during CDFSL. Additional results can be found in Supp. Figs.~\ref{attnn} and \ref{cam-sup}. Quantitative analysis in Fig.~\ref{alignment score}b also confirms that our method 
improves local alignment scores. 

\subsection{Interpretability}
\begin{figure}[thbp]
\vskip -0.1cm   
\centering
\includegraphics[width=0.93\columnwidth]
{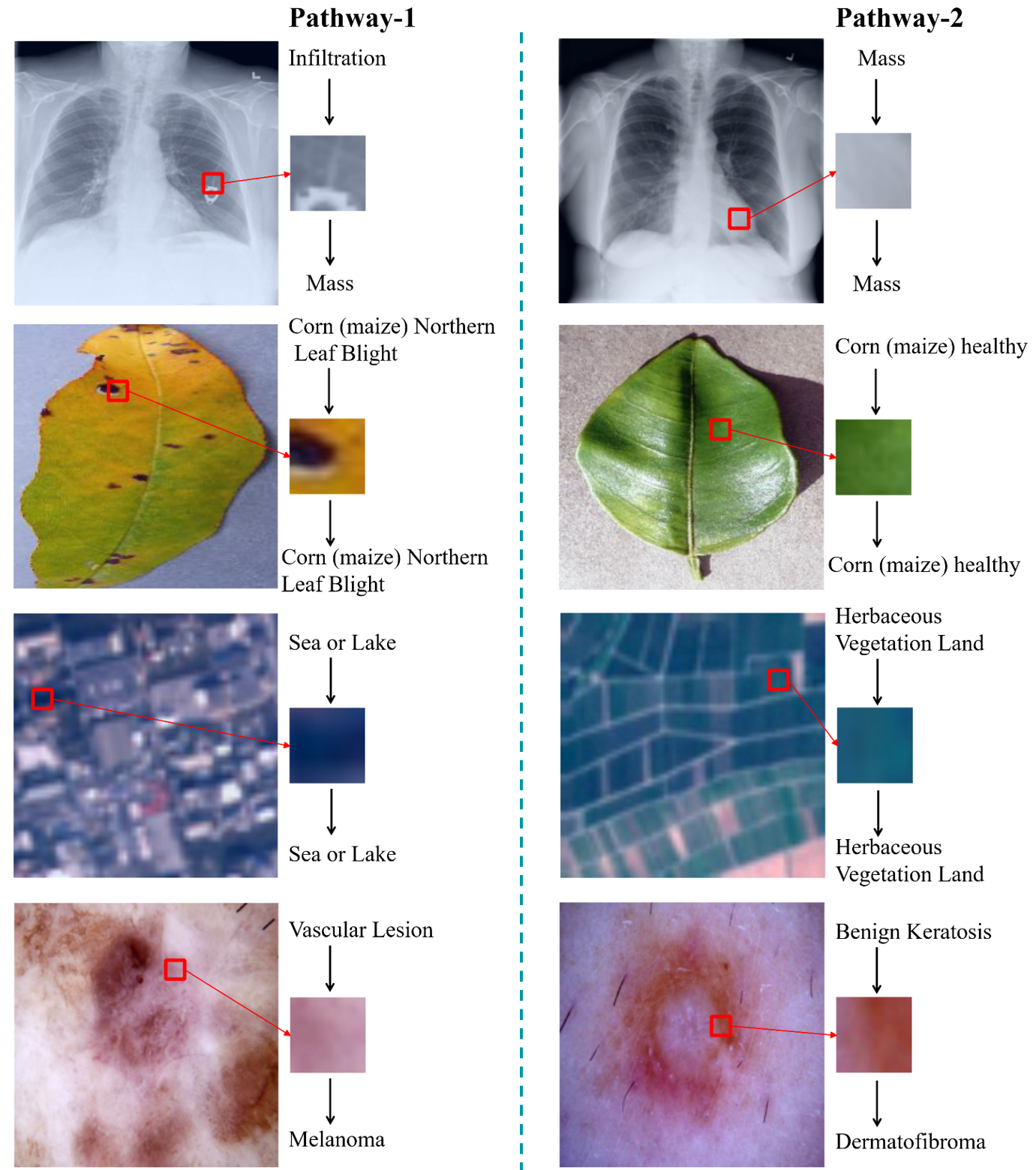} \vspace{-0.2cm}
\caption{The T-I-T pathway indicates that the model can focus on disease-relevant regions and capture fine-grained semantic relationships between different diseases, providing valuable interpretability for medical analysis even when the reconstructed text does not fully match the initial label. }
\label{t2t}
\vspace{-0.2cm}
\end{figure}

\vspace{-0.1cm}
\noindent \textbf{T-I-T cycle pathway}. 
To  validate the interpretability of our approach, we show the
T-I-T cycle consistency pathway, as illustrated in Figure~\ref{t2t}. 
Taking the ChestX-ray image in pathway-1  as an example, the initial text label is ``Infiltration". Through the T-I-T pathway, the model identifies a local region of interest (highlighted by the red box) and maps this patch back to the text space, resulting in the reconstructed text ``Mass". Even though the reconstructed text does not exactly match the initial label, this inconsistency itself provides valuable interpretability. On one hand, it demonstrates that the model is able to focus on disease-relevant regions and detect potential abnormal structures within the image. On the other hand, in medical imaging, different pathological types such as ``Infiltration" and ``Mass" may share similar visual characteristics on chest X-rays, making such semantic deviations in patch-level reconstruction understandable. This phenomenon indicates that the model not only localizes abnormal areas but also captures the fine-grained semantic relationships between different diseases, 
providing a valuable reference for clinicians in their further analysis and diagnosis.

\begin{figure}[htbp]
\centering
\includegraphics[width=0.93\columnwidth]{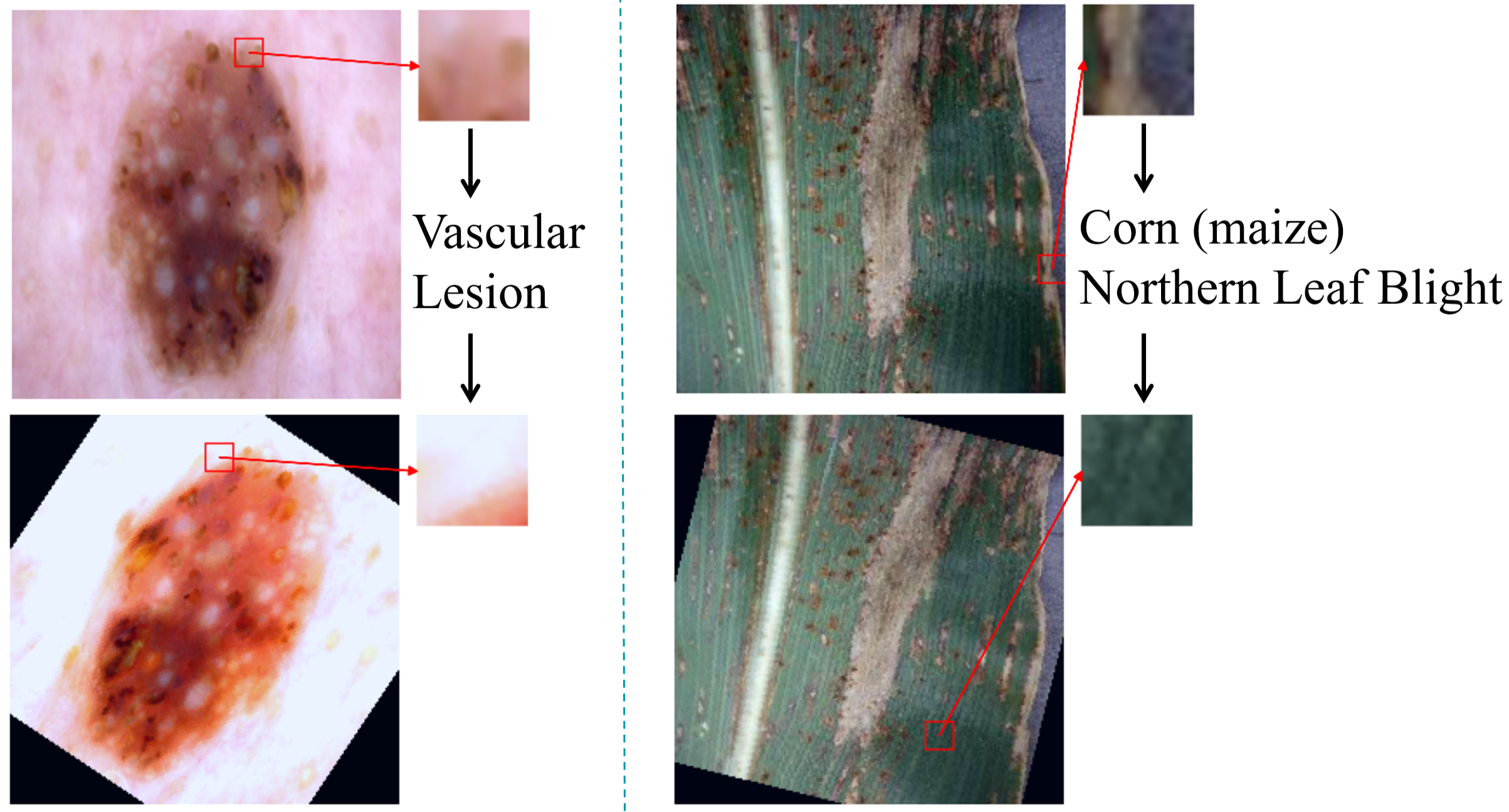}\vspace{-0.2cm} 
\caption{ 
The I-T-I cycle 
pathway demonstrates how the model extracts and retains essential 
semantics in the text domain, which are then used to re-focus on meaningful regions within the image's  augmented feature space.
}
\label{i2i}
\vspace{-0.45cm} 
\end{figure}

\vspace{0.1cm}
\noindent\textbf{I-T-I cycle pathway}.
Figure~\ref{i2i} depicts the I-T-I cycle consistency pathway (Due to space constraints, we present more results in the 
Supplementary Material). 
It
shows 
that the model can extract and retain essential semantics in the text domain, and that these semantics can, in turn, be used to re-focus on meaningful regions within the image. 

\vspace{0.1cm}
\noindent\textbf{Activation Map and Semantic Anchors}.
Our model’s activation maps (Fig.~\ref{attn}) precisely highlight 
discriminative fine-grained cues. 
Semantic Anchor’s shrink phase  provides interpretability by visualizing the model's focus (Fig.~\ref{select}).

\begin{table}[b]
\setlength{\abovecaptionskip}{2pt}  
\vspace{-0.4cm}
\caption{Prototype-classification accuracy 
of the 
 vision encoder.  
}
\centering
\small
\resizebox{0.4\textwidth}{!}{
\setlength{\tabcolsep}{1.1mm}
\begin{tabular}{cccccc}
\toprule
  & ISIC & ChestX & EuroSAT &  CropDiseases  \\
\midrule
CLIP-LoRA & 49.83 & 23.79 & 93.0 &  95.85  \\
+ Ours &  50.68 & 24.51 & 93.64 & 96.24   \\
\bottomrule
\end{tabular}
}  
\label{tab:proto}
\end{table}

\vspace{-0.1cm}
\subsection{Beneficial to Discriminability}
\vspace{-0.1cm}
Notably, the above inconsistency  is not a model failure 
but rather evidence 
that our method successfully learns transferable, fine-grained visual concepts shared across semantically related categories. By identifying these common concepts, the model is compelled to learn more nuanced representations that capture the subtle distinctions between classes, thereby improving feature discriminability. 
To verify discriminability, we conducted prototype classification 
using only the 
vision encoder. Tab.~\ref{tab:proto} shows 
consistent improvements 
after applying our cycle-consistent regularizer. 
The enhanced feature discriminability is also visually confirmed by the t-SNE plots in Supplementary Figure~\ref{fig:tsne}. 


\vspace{-0.1cm}
\section{Conclusion}
\vspace{-0.1cm}
We proposed CC-CDFSL, a self-supervised 
regularization framework that enhances local feature alignment in CLIP-based 
CDFSL. By introducing the T-I-T, I-T-I cycle, and complementary Semantic Anchor module, our method effectively captures fine-grained semantics, improving robustness and interpretability. Extensive experiments on 
various 
benchmark datasets demonstrate superior performance.

\section*{Acknowledgments}

This work is supported by the National Natural Science Foundation of China under grants 62206102; the National Key Research and Development Program of China under grant 2024YFC3307900; the National Natural Science Foundation of China under grants 62436003, 62376103 and 62302184; Major Science and Technology Project of Hubei Province under grant 2025BAB011 and 2024BAA008; Hubei Science and Technology Talent Service Project under grant 2024DJC078; and Ant Group through CCF-Ant Research Fund. The computation is completed in the HPC Platform of Huazhong University of Science and Technology.

%% file: sec/X_suppl.tex
\clearpage
\setcounter{page}{1}
\maketitlesupplementary


\section{Detailed Dataset Description}
\begin{figure}[ht]
\centering
\includegraphics[width=0.95\columnwidth]{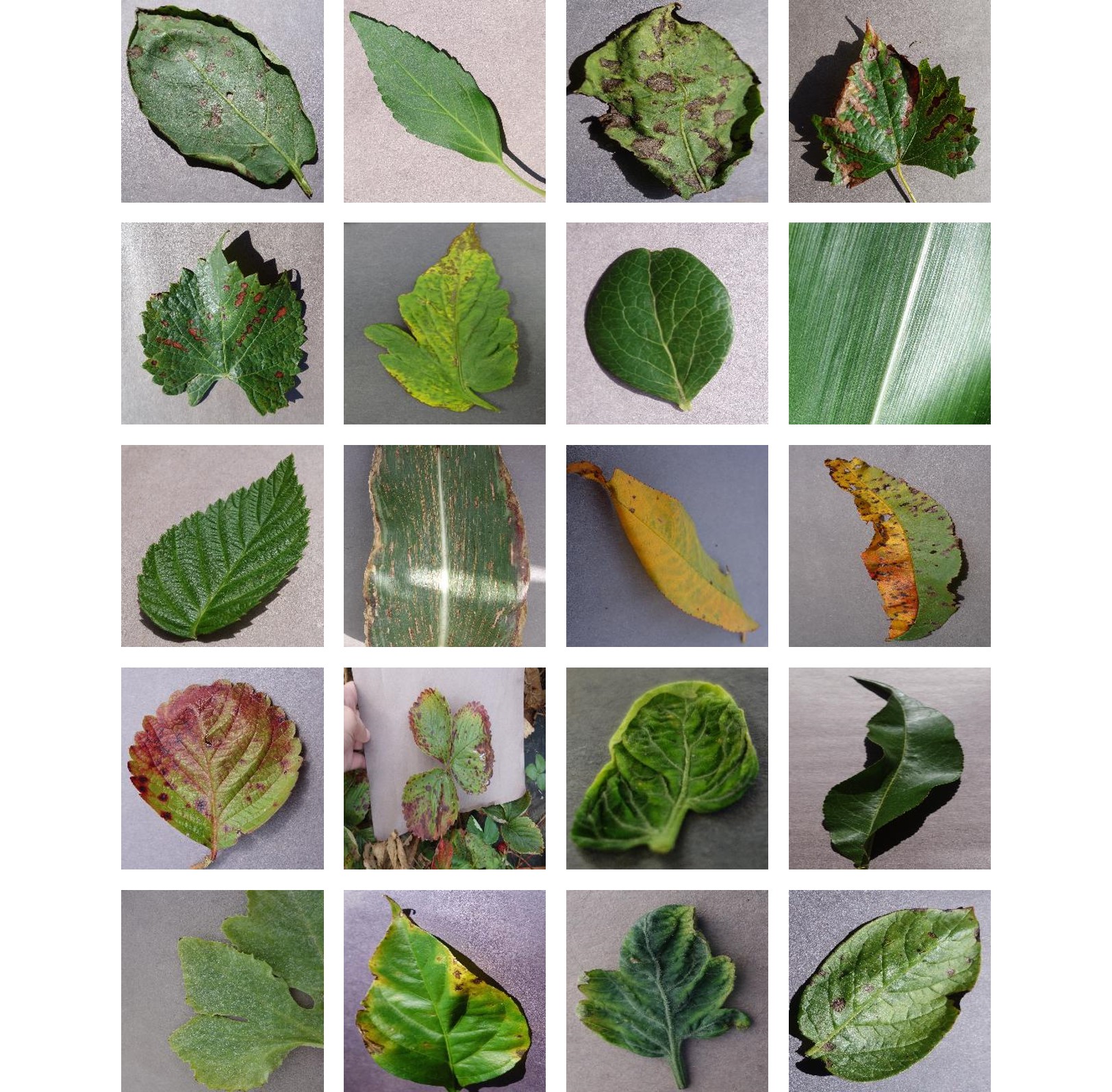}
\caption{Samples from CropDiseases.}
\label{fig:crop}
\vspace{-0.3cm}
\end{figure}

Our experimental setup follows the BSCD-FSL~\cite{guo2020broader} benchmark, addressing the challenge of significant distributional shifts across four distinct target domain datasets. Detailed information on these datasets is provided below:

  \textbf{CropDiseases}~\cite{mohanty2016using} is a dataset including 54,306 images of 14 crop species (Apple, Blueberry, Cherry, Corn, Grape, Orange, Peach, Bell Pepper, Potato, Raspberry, Soybean, Squash, Strawberry, and Tomato) with 26 diseases (or healthy). The samples of this dataset are listed in Fig.~\ref{fig:crop}. CropDiseases images are natural images, but are very specialized 
  (specific to the agriculture industry), so the domain gap here is larger than in the previous cross-domain setting~\cite{DBLP:conf/iclr/TsengLH020}. 

\begin{figure}[ht]
\centering
\includegraphics[width=0.95\columnwidth]{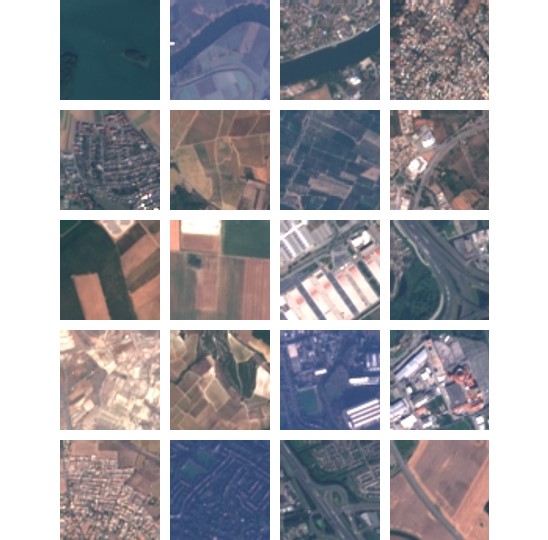}
\caption{Samples from EuroSAT.}
\label{fig:eurosat}
\vspace{-0.5cm}
\end{figure}

  \textbf{EuroSAT} \cite{helber2019eurosat} is a dataset for land use and land cover classification. EuroSAT based on Sentinel-2 satellite imagery, covers 13 spectral bands and consists of 10 categories including Industrial Buildings, Residential Buildings, Annual Crop, Permanent Crop, River, Sea \& Lake, Herbaceous Vegetation, Highway, Pasture and Forest, with a total of 27,000 annotated and geographically referenced images. Compared to CropDiseases, EuroSAT images are less similar to \textit{mini}Imagenet (source domain dataset) as they have lost perspective distortion, but are still color images of natural scenes.
   The samples of this dataset are listed in Fig.~\ref{fig:eurosat}.

\begin{figure}[ht]
\centering
\includegraphics[width=0.95\columnwidth]{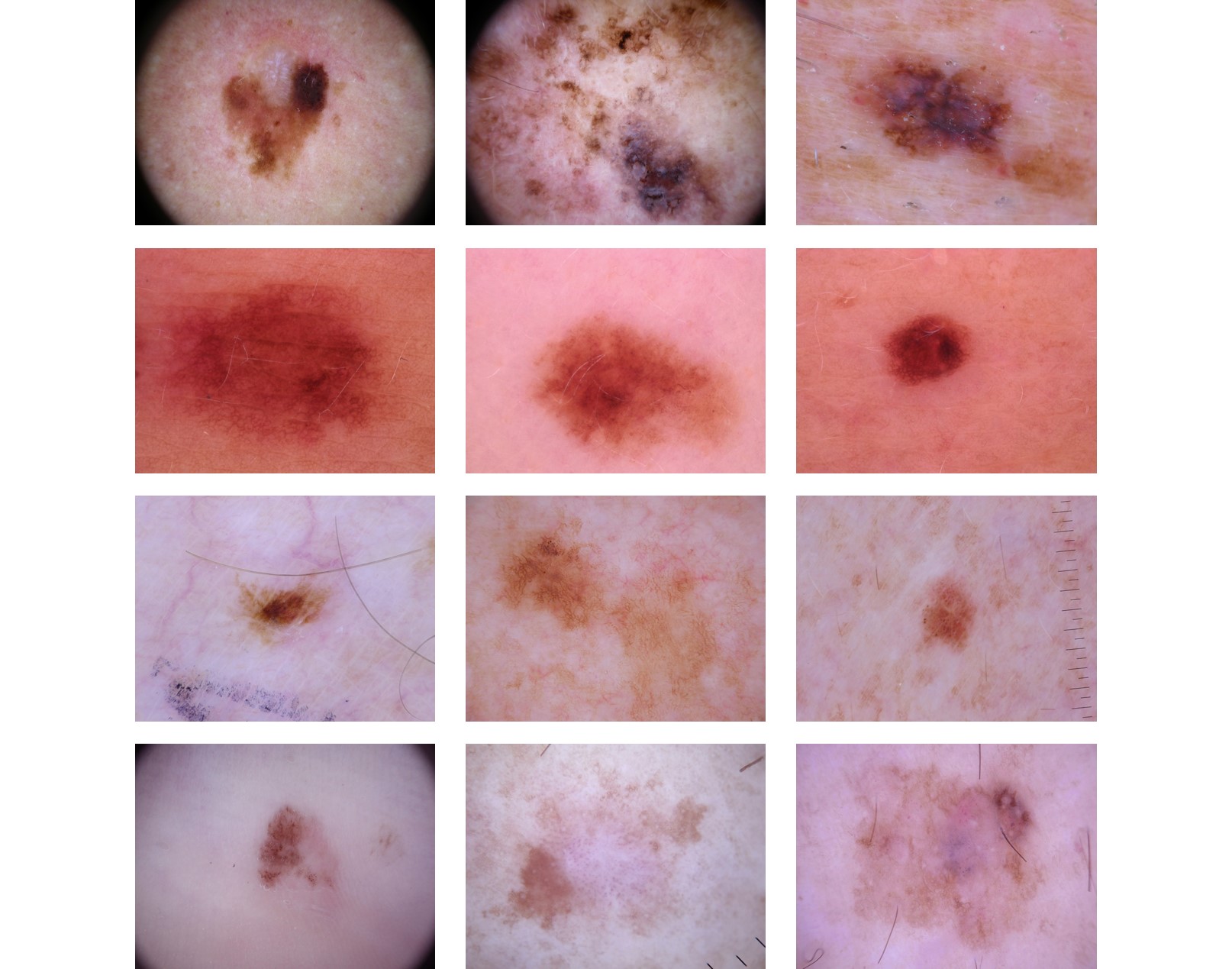}
\caption{Samples from ISIC2018.}
\label{fig:isic}
\vspace{-0.3cm}
\end{figure}

The   \textbf{ISIC2018} \cite{codella2019skin} dataset was published by the International Skin Imaging Collaboration (ISIC) as a large-scale dataset of dermoscopy images containing
10,015 images of seven skin injury types (melanoma, melanocytic nevus, basal cell carcinoma, actinic keratosis, benign keratosis, dermatofibroma, or a vascular lesion). ISIC2018 images are even less similar to \textit{mini}Imagenet as they have lost perspective distortion and no longer represent natural scenes.
The samples of this dataset are listed in Fig. \ref{fig:isic}.

\begin{figure}[ht]
\centering
\includegraphics[width=0.95\columnwidth]{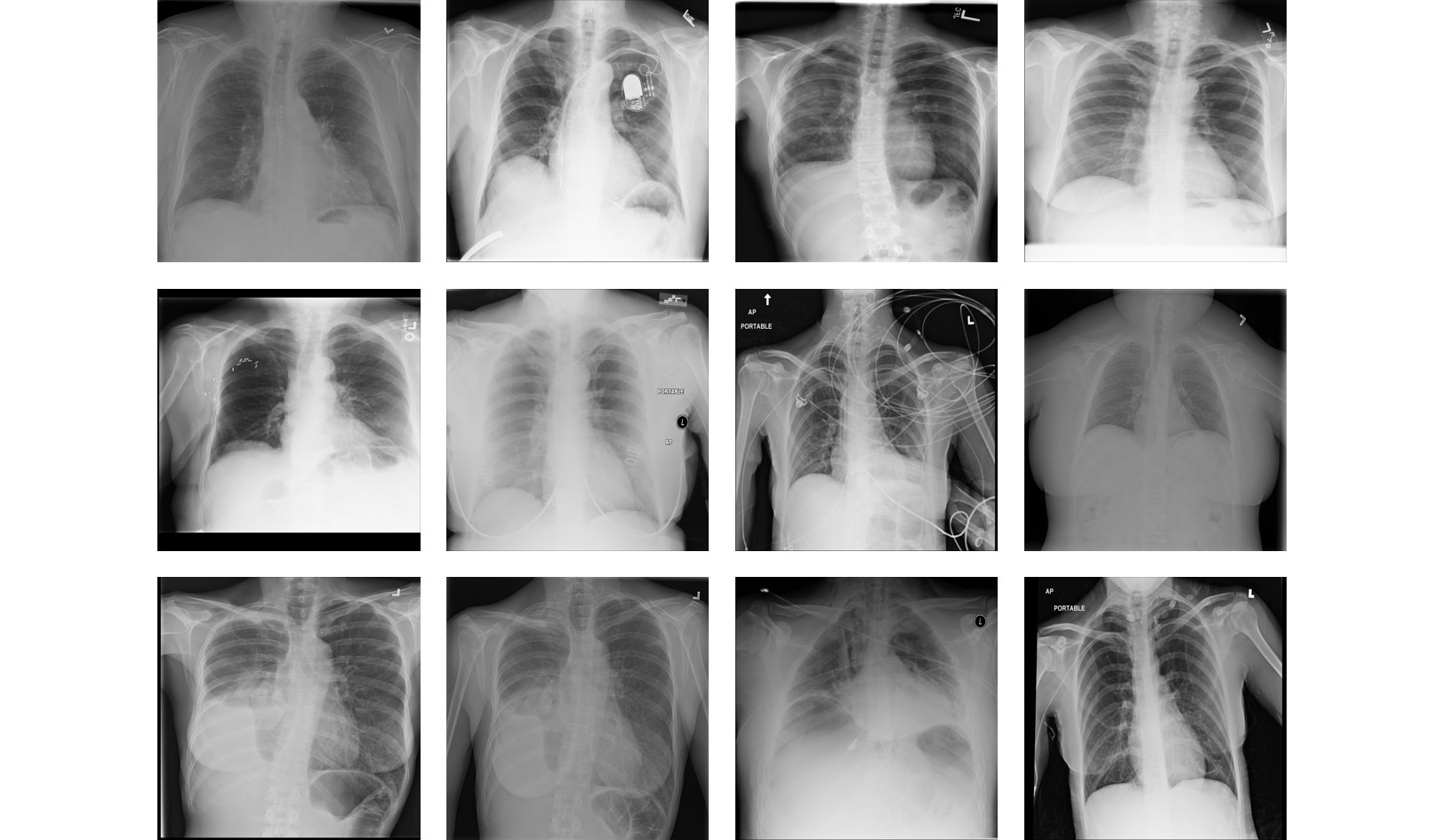}
\caption{Samples from ChestX.}
\label{fig:chestx}
\vspace{-0.5cm}
\end{figure}

ChestX-ray14 is the largest lung X-ray database to date, which contains more than 100,000 pre-X-ray views for 14 lung diseases. Categories 1 to 14 correspond to 14 lung diseases, and category 15 indicates no disease. \cite{wang2017chestx} studied the images of eight diseases in this database and constructed the ChestX-ray8 dataset, which comprises 108,948 frontal view X-ray images of 32,717 unique patients with the text-mined eight disease image labels (where each image can have multi-labels) from the associated radiological reports using natural language processing. In this work, we use   \textbf{ChestX-ray8} for cross-domain testing, consistent with \cite{guo2020broader}. ChestX is the most dissimilar to \textit{mini}Imagenet across the four target domains as 
its images 
have lost perspective distortion, do not represent natural scenes, and have lost 2 color channels.
The samples of this dataset are listed in Fig.~\ref{fig:chestx}.
 \vspace{-0.1cm}




%

\section{Applying Our Method to  other ViT Variants}
Table~\ref{tab:vit} presents a comparative analysis of our method against the CLIP-LoRA baseline across other ViT architectures, specifically ViT-B/32, 
and ViT-L/14, under the 1-shot setting. As observed, our method consistently demonstrates significant performance improvements over the baseline across all tested ViT variants and target domains. 

\begin{table}[htbp] 
\setlength{\abovecaptionskip}{2pt}  
\caption{Comparison with ViT-based methods in 1-shot.}
\small
\setlength{\tabcolsep}{1mm} 
\centering
\begin{tabular}{@{}ccccccc@{}}
\toprule
Backbone & Method & 
ChestX & ISIC & EuroSAT & Crop. & Ave. \\
\midrule

\multirow{2}{*}{ViT-B/32} 
& Baseline & 21.23 & 36.32 & 79.79 & 81.13 & 54.62 \\ 
& + Ours  &    21.36 & 38.26 & 82.88 & 83.98 & 56.62 \\ 
\midrule


\multirow{2}{*}{ViT-L/14} 
& Baseline & 22.90 & 38.45 & 85.83 & 89.83 & 59.25 \\ 
& + Ours   & 23.19 & 38.72 & 90.11   & 90.77 & 60.70 \\ 
\bottomrule

\end{tabular}
\label{tab:vit}
\vspace{-0.3cm}
\end{table}

\section{Experimental Data of Figure 2}


Figure~\ref{alignment score} in the main paper quantifies the hypothesis that the domain gap and scarce training data in CDFSL hurt patch-level local alignment more severely than global alignment. Table~\ref{tab:align} reports these alignment scores for both global (CLS-token) and local (patch-level) representations of the baseline, CLIP-LoRA \cite{zanella2024low}. 
As observed, all cross-domain datasets exhibit lower local alignment scores compared to miniImageNet, with the exception of EuroSAT, whose images are dominated by foreground regions, 
 confirming the baseline's inability to capture fine-grained cues under data scarcity. 
To further validate the efficacy of the proposed CC-CDFSL framework, we compare the local alignment scores before and after applying our method. Table~\ref{tab:tar} summarizes the results.
Consistently, our method significantly improves these scores, demonstrating that the cycle-consistency and semantic anchor mechanisms successfully 
enhance 
patch-level semantics alignment under domain shift without additional annotations.

\begin{table}[htbp]
\setlength{\abovecaptionskip}{2pt}  
\caption{Alignment Scores of Global and Local Features.}
    \centering
    \small
    \begin{tabular}{@{}cccc@{}}
        \toprule
        Dataset & Global Feature & Local Features & Far Domain \\ 
        \midrule
        miniImageNet   & 0.7056 & 0.3821 & \ding{55} \\
        ChestX         & 0.6768 & 0.2391 & \checkmark  \\
        ISIC2018       & 0.7261 & 0.2360 & \checkmark  \\
        EuroSAT        & 0.6768 & 0.3845 & \checkmark  \\
        CropDiseases   & 0.7012 & 0.3811 & \checkmark  \\
        \bottomrule
    \end{tabular}
    \label{tab:align}
\vspace{-0.3cm}
\end{table}

\begin{table}[htbp]
\setlength{\abovecaptionskip}{2pt}  
\caption{Comparison of Local Feature Alignment Scores.
} 
\centering
\small
\setlength{\tabcolsep}{1.1mm} 
\begin{tabular}{lccccc}
\toprule
Method & ChestX & ISIC & EuroSAT & Crop. & Ave. \\
\midrule
Baseline  & 0.2391 & 0.2360 & 0.3845 & 0.3811 & 0.3102 \\
+ Ours      & 0.3293 & 0.2661 & 0.3796 & 0.3877 &  0.3407 \\
\bottomrule
\end{tabular}
\label{tab:tar}
 \vspace{-0.3cm}
\end{table}

\begin{table*}[tbp]
  \centering
  \small
  \setlength{\abovecaptionskip}{2pt}  
  \caption{Base-to-new generalization on 11 datasets. Ours (CC-CDFSL) consistently improves new-class accuracy and harmonic mean (HM) over MaPLe, validating the cross-task generality of cycle-consistent patch-level alignment.}
\begin{subfigure}[b]{0.32 \textwidth}
    \setlength{\abovecaptionskip}{1pt}
    \setlength{\belowcaptionskip}{1pt}
    \caption{Average over 11 datasets.}
    \begin{tabular}{lccc}
        \toprule
        & Base & New & HM \\
        \midrule
        MaPLe & 82.19 & 74.55 & 78.18 \\
        + Ours & 82.26 & 75.75 & 78.87 \\
        \bottomrule
    \end{tabular}
\end{subfigure}
\begin{subfigure}[b]{0.32	\textwidth}
    \setlength{\abovecaptionskip}{1pt}
    \setlength{\belowcaptionskip}{1pt}
    \caption{ImageNet.}
    \begin{tabular}{lccc}
        \toprule
        & Base & New & HM \\
        \midrule
        MaPLe & 75.57 & 70.88 & 73.15 \\
        + Ours & 75.59 & 71.10 & 73.28 \\
        \bottomrule
    \end{tabular}
\end{subfigure}
\begin{subfigure}[b]{0.32 \textwidth}
    \setlength{\abovecaptionskip}{1pt}
    \setlength{\belowcaptionskip}{1pt}
    \caption{Caltech101.}
 
    \begin{tabular}{lccc}
        \toprule
        & Base & New & HM \\
        \midrule
        MaPLe & 98.02 & 94.43 & 96.19 \\
        + Ours & 97.91 & 95.49 & 96.68 \\
        \bottomrule
    \end{tabular}
\end{subfigure}

\begin{subfigure}[b]{0.32	\textwidth}
    \setlength{\abovecaptionskip}{1pt}
    \setlength{\belowcaptionskip}{1pt}
    \caption{OxfordPets.}
 
    \begin{tabular}{lccc}
        \toprule
        & Base & New & HM \\
        \midrule
        MaPLe & 95.64 & 97.80 & 96.71 \\
        + Ours & 95.82 & 97.99 & 96.89 \\
        \bottomrule
    \end{tabular}
    
\end{subfigure}
\begin{subfigure}[b]{0.32	\textwidth}
    \setlength{\abovecaptionskip}{1pt}
    \setlength{\belowcaptionskip}{1pt}
    \caption{StanfordCars.}
 
    \begin{tabular}{lccc}
        \toprule
        & Base & New & HM \\
        \midrule
        MaPLe & 72.56 & 73.80 & 73.17 \\
        + Ours & 72.66 & 74.84 & 73.73 \\
        \bottomrule
    \end{tabular}
    
\end{subfigure}
\begin{subfigure}[b]{0.32	\textwidth}
    \setlength{\abovecaptionskip}{1pt}
    \setlength{\belowcaptionskip}{1pt}
    \caption{Flowers102.}
 
    \begin{tabular}{lccc}
        \toprule
        & Base & New & HM \\
        \midrule
        MaPLe & 96.11 & 72.44 & 82.61 \\
        + Ours & 96.07 & 73.90 & 83.19 \\
        \bottomrule
    \end{tabular}
    
\end{subfigure}

\begin{subfigure}[b]{0.32	\textwidth}
    \setlength{\abovecaptionskip}{1pt}
    \setlength{\belowcaptionskip}{1pt}
    \caption{Food101.}
 
    \begin{tabular}{lccc}
        \toprule
        & Base & New & HM \\
        \midrule
        MaPLe & 90.75 & 91.84 & 91.29 \\
        + Ours & 90.83 & 92.00 & 91.41 \\
        \bottomrule
    \end{tabular}
    
\end{subfigure}
\begin{subfigure}[b]{0.32	\textwidth}
    \setlength{\abovecaptionskip}{1pt}
    \setlength{\belowcaptionskip}{1pt}
    \caption{FGVCAircraft.}
 
    \begin{tabular}{lccc}
        \toprule
        & Base & New & HM \\
        \midrule
        MaPLe & 38.34 & 34.59 & 36.37 \\
        + Ours & 38.00 & 36.01 & 36.98 \\
        \bottomrule
    \end{tabular}
    
\end{subfigure}
\begin{subfigure}[b]{0.32	\textwidth}
    \setlength{\abovecaptionskip}{1pt}
    \setlength{\belowcaptionskip}{1pt}
    \caption{SUN397.}
 
    \begin{tabular}{lccc}
        \toprule
        & Base & New & HM \\
        \midrule
        MaPLe & 80.92 & 78.21 & 79.54 \\
        + Ours & 81.04 & 79.08 & 80.05 \\
        \bottomrule
    \end{tabular}
    
\end{subfigure}

\begin{subfigure}[b]{0.32	\textwidth}
    \setlength{\abovecaptionskip}{1pt}
    \setlength{\belowcaptionskip}{1pt}
    \caption{DTD.}
 
    \begin{tabular}{lccc}
        \toprule
        & Base & New & HM \\
        \midrule
        MaPLe & 79.86 & 59.54 & 68.28 \\
        + Ours & 79.90 & 62.28 & 70.00 \\
        \bottomrule
    \end{tabular}
    
\end{subfigure}
\begin{subfigure}[b]{0.32	\textwidth}
    \setlength{\abovecaptionskip}{1pt}
    \setlength{\belowcaptionskip}{1pt}
    \caption{EuroSAT.}
    \begin{tabular}{lccc}
        \toprule
        & Base & New & HM \\
        \midrule
        MaPLe & 92.66 & 68.39 & 78.70 \\
        + Ours & 93.57 & 71.98 & 81.37 \\
        \bottomrule
    \end{tabular}
    
\end{subfigure}
\begin{subfigure}[b]{0.32	\textwidth}
    \setlength{\abovecaptionskip}{1pt}
    \setlength{\belowcaptionskip}{1pt}
    \caption{UCF101.}
    \begin{tabular}{lccc}
        \toprule
        & Base & New & HM \\
        \midrule
        MaPLe & 83.66 & 78.04 & 80.75 \\
        + Ours & 83.51 & 79.07 & 81.23 \\
        \bottomrule
    \end{tabular}
\end{subfigure}
\label{tab:compare}
\vspace{-0.3cm}
\end{table*}

\section{Generalization to Base-to-New 
 Setting 
}

 The  base-to-new generalization task refers to a scenario where a dataset is evenly divided by class into non-overlapping base classes and new classes. After a model is trained on a few-shot dataset from the base classes, its generalization ability is tested on the new classes. The harmonic mean (HM) of the classification accuracies on both is used to evaluate the overall performance:
\begin{equation}
\label{eq:hm}
\text{HM} = \frac{2}{\frac{1}{\text{Base}} + \frac{1}{\text{New}}} = \frac{2 \cdot \text{Base} \cdot \text{New}}{\text{Base} + \text{New}}
\end{equation}

As shown in Table~\ref{tab:compare}, CC-CDFSL consistently outperforms MaPLe \cite{khattakMaPLe} across all datasets in terms of new-class accuracy and harmonic mean, with 
 marginal improvements
on base classes. 


\subsection{Better Class Separation} 
\begin{figure}[htbp]
    \centering
    \begin{tabular}{@{}ccc@{}}
        \includegraphics[width=0.14\textwidth]{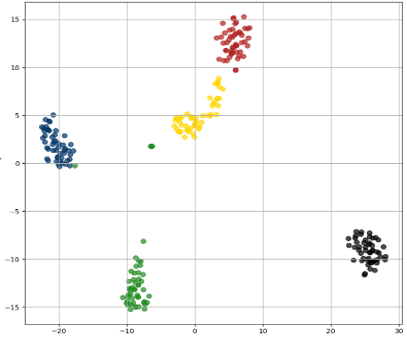} & 
        \includegraphics[width=0.14\textwidth]{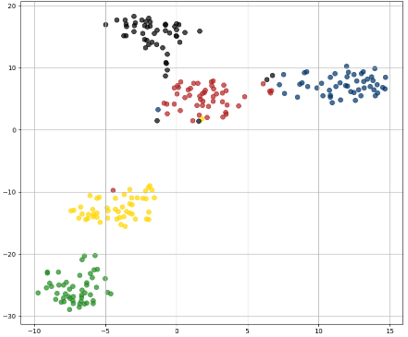} & 
        \includegraphics[width=0.14\textwidth]{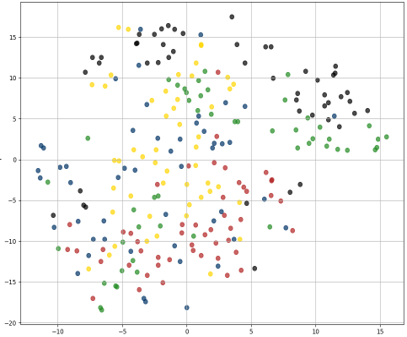} \\
        \includegraphics[width=0.14\textwidth]{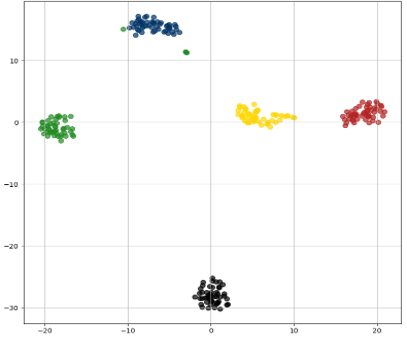} & 
        \includegraphics[width=0.14\textwidth]{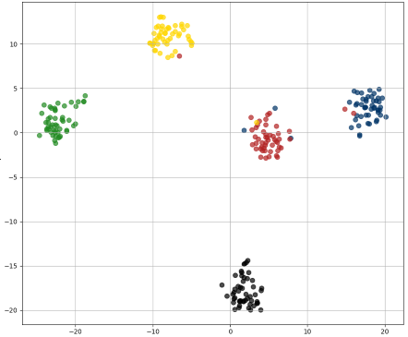} & 
        \includegraphics[width=0.14\textwidth]{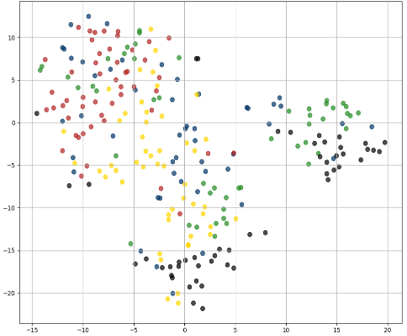} \\
        \small CropDiseases & \small EuroSAT & \small ISIC \\
    \end{tabular}
    \caption{t-SNE visualization of feature distributions on different datasets. The first row shows results from CLIP-LoRA. The second row shows results from our proposed method.  Different colors denote different classes.
    }
    \label{fig:tsne}
\vspace{-0.3cm}
\end{figure}

As illustrated in the Figure \ref{fig:tsne}, features extracted by CLIP-LoRA tend to form less compact and more overlapping clusters, with different classes not well separated in the embedding space. In contrast, our method produces more distinct and compact clusters for each class, resulting in clearer class boundaries and reduced intra-class variance. This indicates that our approach achieves better class separation compared to the baseline.
For ChestX, t-SNE embeddings are omitted: the grayscale chest X-rays exhibit minuscule lesion regions that yield highly overlapping clusters, rendering any inter-method distinctions imperceptible.

\vspace{-0.2cm}
\section{Visualization}

\subsection{Improved Focus on Relevant Semantics} 
\begin{figure}[htbp]
\centering
\includegraphics[width=0.98\columnwidth]
{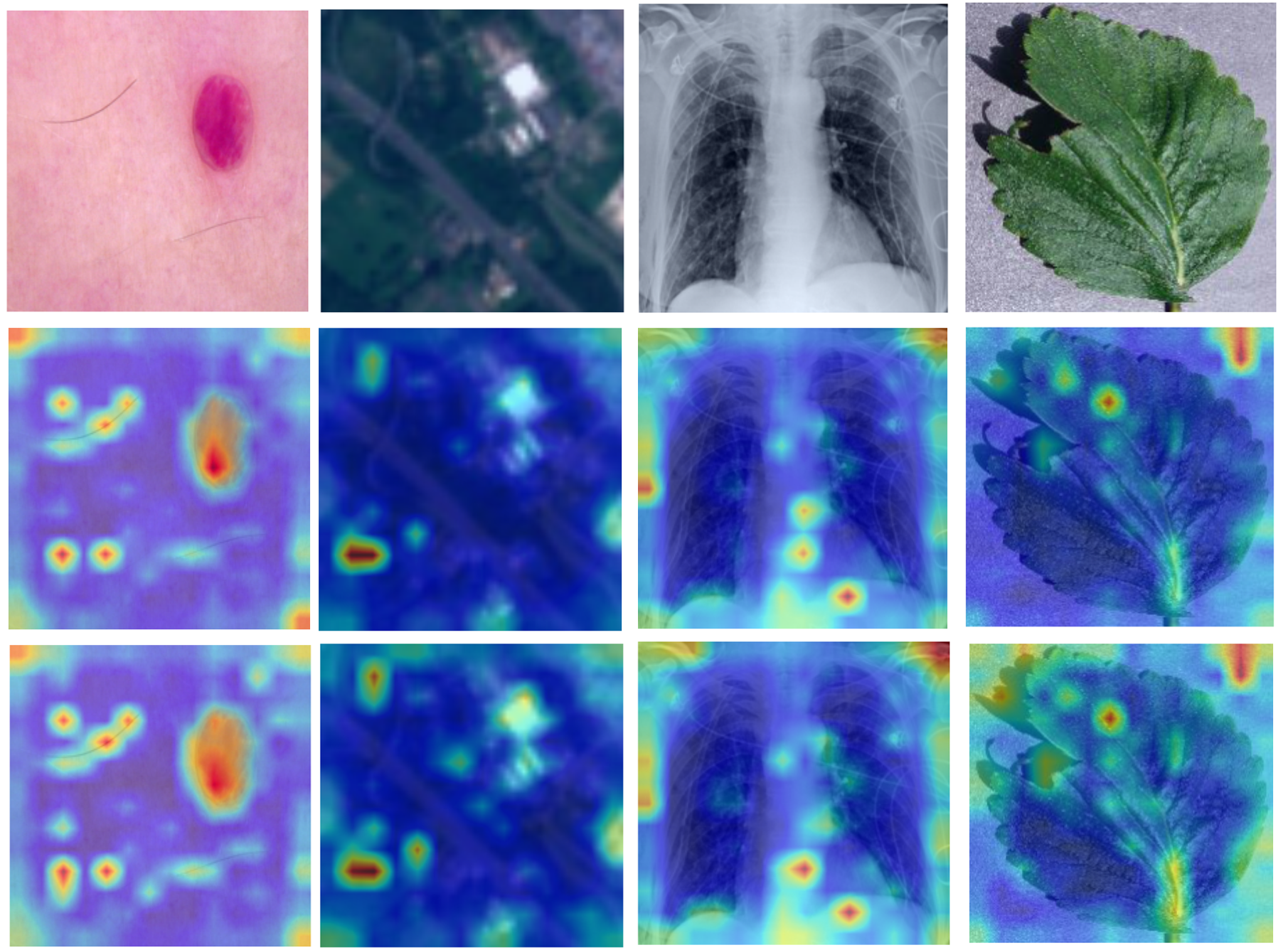} 
\begin{tabular}{cccc}
 \small \makebox[0.2 \columnwidth][c]{ISIC} & \small \makebox[0.2\columnwidth][c]{EuroSAT} & \small \makebox[0.2\columnwidth][c]{ChestX} & \small 
 \makebox[0.2\columnwidth][c]{CropDiseases} \\
\end{tabular}
\caption{
Row 2 and Row 3 display the attention maps of the baseline and CC-CDFSL, respectively, showing that the latter localizes critical regions more precisely.
}  
\label{attnn} 
\vspace{-0.3cm} 
\end{figure}

\begin{figure}[htbp]
\centering
\includegraphics[width=0.98\columnwidth]{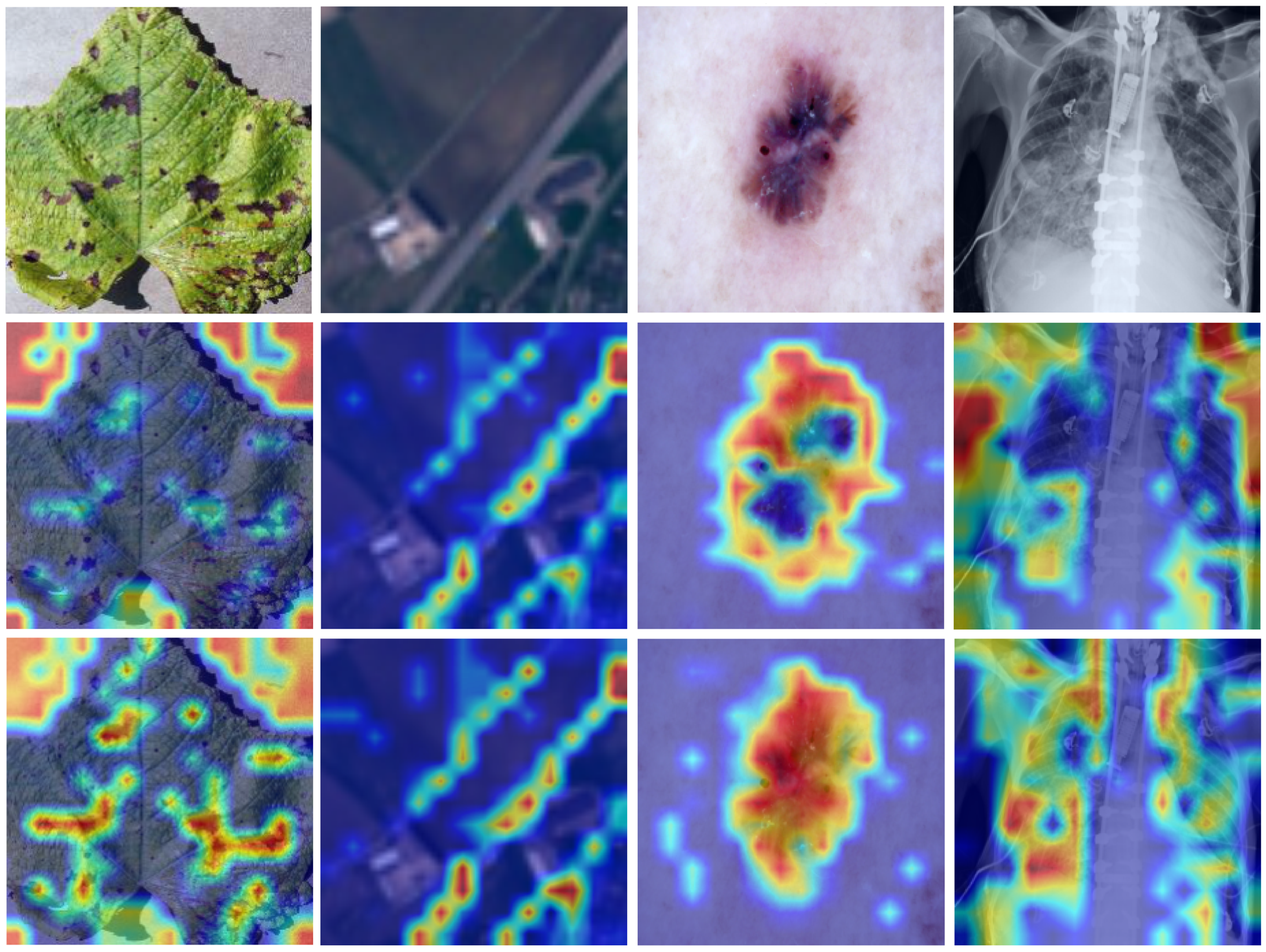} \\
\begin{tabular}{cccc}
 \small \makebox[0.2 \columnwidth][c]{CropDiseases} & \small \makebox[0.21\columnwidth][c]{EuroSAT} & \small \makebox[0.2\columnwidth][c]{ISIC} & \small 
 \makebox[0.2\columnwidth][c]{ChestX} \\
\end{tabular}
\caption{The heatmap for CLIP-LoRA (the second row) and our CC-CDFSL (the third row) in four target domains.}  
\label{cam-sup}
\vspace{-0.3cm}
\end{figure}

 Figure \ref{attnn} and Figure \ref{cam-sup}  respectively present the attention maps and Grad-CAM \cite{DBLP:conf/iccv/SelvarajuCDVPB17} heatmaps produced by the baseline model and our proposed method across four datasets.
This demonstrates that our method enhances the model's ability to focus on critical features in  
cross-domain few-shot learning tasks and simultaneously improves local alignment between local 
visual patches and textual semantics.

\begin{figure}[thbp]
\centering
\includegraphics[width=0.98\columnwidth]{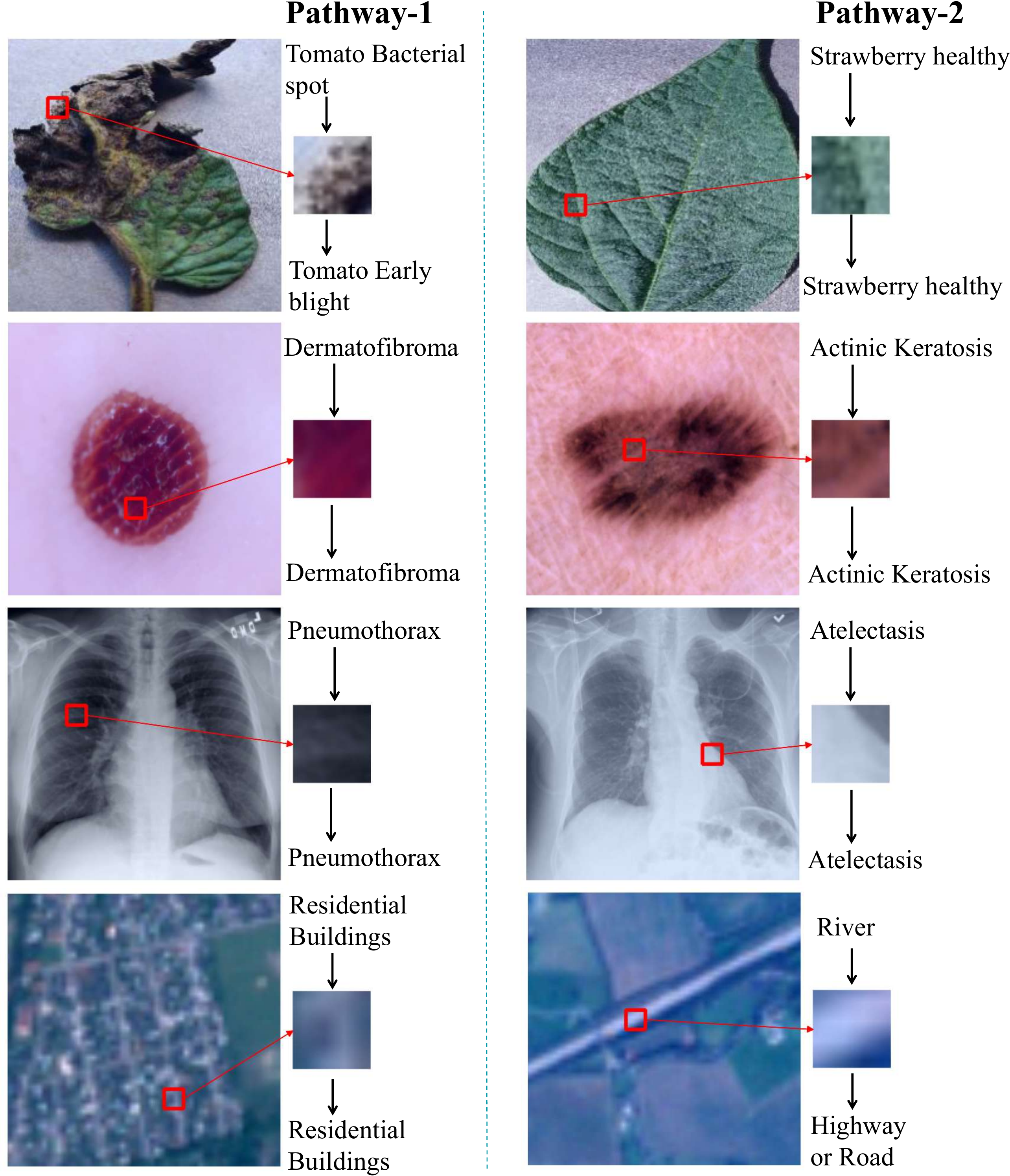} 
\caption{Illustration of the Text-to-Image-patch-to-Text (T-I-T) cycle consistency pathway. Each  specific class text  is first used to locate the most semantically relevant local patch within the image. The selected image patch is then mapped back to the text space in an attempt to reconstruct the original class label. And the patches in the figure are resized to a resolution of 64×64  for better presentation.}
\label{t2tt}
 \vspace{-0.5cm}
\end{figure}    

\begin{figure}[ht]
\centering
\includegraphics[width=0.98\columnwidth]{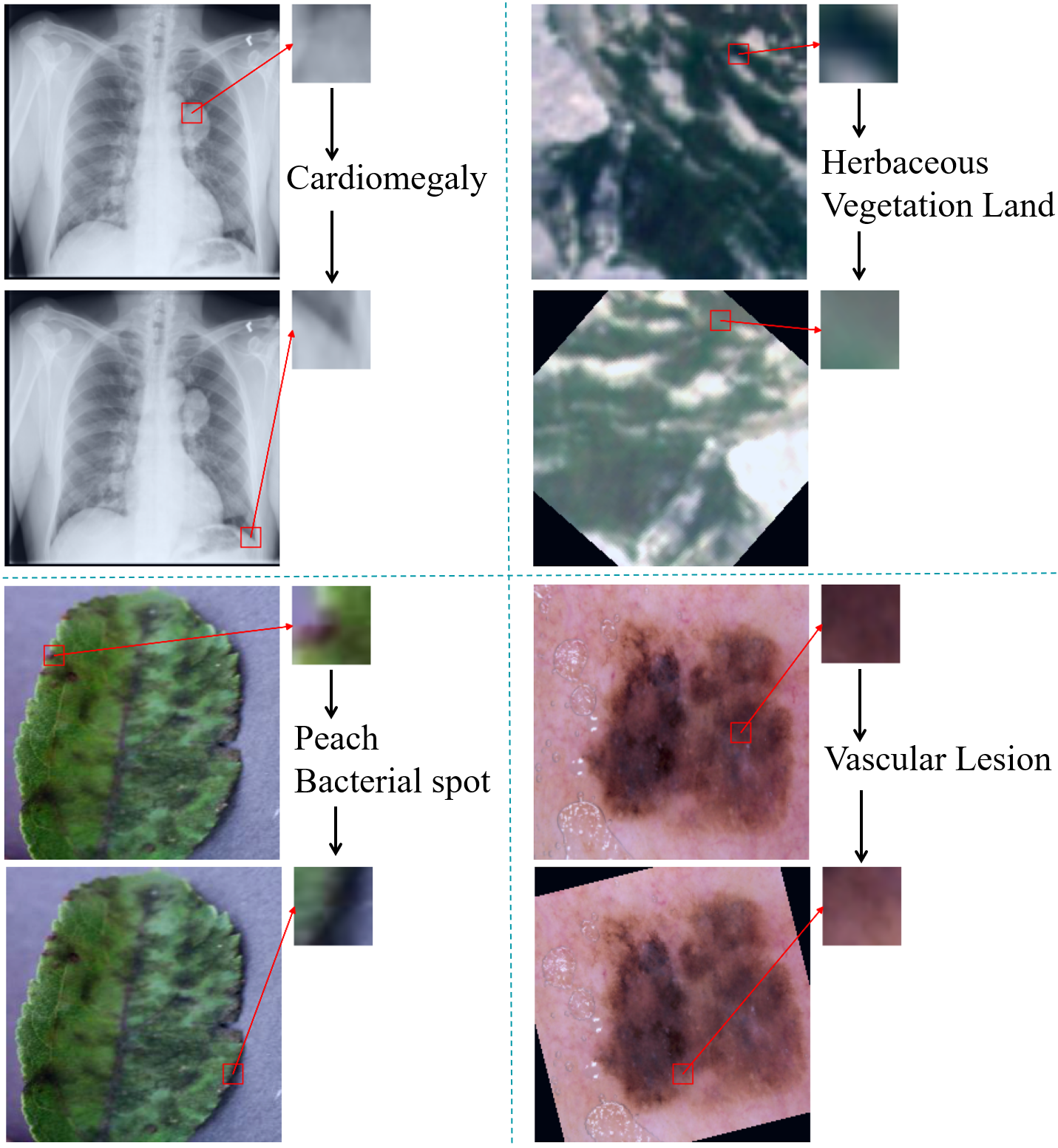} 
\caption{Additional visualizations of the Image-to-Text-to-Image (I-T-I) cycle consistency pathway. Each pathway shows a selected patch within the initial image, the most similar text derived from this patch, and a semantically  re-focused   patch within its corresponding  image in the augmented image space.} 
\label{i2ii}
\end{figure}    


\section{Comparison with SOTA  CDFSL Methods} 

We also evaluate our  CC-CDFSL framework in comparison with the most competitive state-of-the-art (SOTA)  CDFSL methods, covering a range of settings such as different backbones, the use of source datasets, and whether fine-tuning is performed on the target domain (FT).
 MEM-FS \cite{DBLP:journals/tip/WalshOS23}, StyleAdv \cite{DBLP:conf/cvpr/FuXFJ23}, 
FLoR \cite{DBLP:conf/cvpr/ZouLH0024},  DAMIM \cite{DBLP:conf/aaai/MaZ0025}, AttnTemp \cite{zou2024attention}, CD-CLS \cite{zou2024a}, PMF \cite{DBLP:conf/cvpr/Hu0SKH22}, IM-DCL \cite{DBLP:journals/tip/XuLZFSCL24}, StepSPT \cite{DBLP:journals/corr/abs-2411-10070},  and SeGD-VPT \cite{DBLP:journals/corr/abs-2412-00767}   are introduced as competitors.
Tabs. \ref{tab:5shot} and \ref{tab:1shot} present that our method achieves new SOTA average accuracies of 67.90\% and 58.85\% for the 5-way 5-shot and 5-way 1-shot classification tasks, respectively.
Notably, IM-DCL \cite{DBLP:journals/tip/XuLZFSCL24} 
excels on ChestX, 
which we attribute to its ResNet-based backbone being better suited for 
capturing local features prevalent in medical images.

\begin{table*}[thbp]
\setlength{\abovecaptionskip}{2pt}  
\caption{Comparison with state-of-the-art CDFSL works by the 5-way 5-shot classification.}
\centering
\resizebox{0.9\textwidth}{!}{
\begin{tabular}{lcc*{7}{c}} 
\toprule
Method & Backbone & Mark & Source & Target & CropDiseases & EuroSAT & ISIC & ChestX & Ave.\\
\midrule
MEM-FS & ViT/DINO & TIP-23 & & &93.74&86.49&47.38&26.67&63.57\\
StyleAdv&ViT/DINO & CVPR-23 & \checkmark&- &94.85&88.57&47.73&26.97&64.53\\
FLoR&ViT/DINO & CVPR-24 & \checkmark&- &95.28&90.41&49.52&27.28&65.48\\
DAMIM&ViT/DINO & AAAI-25  & \checkmark&- &95.52&89.50&50.76&27.28&65.77\\
AttnTemp&ViT/DINO & NeurIPS-24 & \checkmark &- &95.53&90.13&53.09&27.72&66.62\\
CD-CLS&ViT/DINO & NeurIPS-24 & \checkmark&\checkmark &96.27&91.53&54.69&27.66&67.54\\
PMF&ViT/DINO & CVPR-22 & \checkmark&\checkmark &92.96&85.98&50.12&27.27&64.08\\
StyleAdv-FT&ViT/DINO &  CVPR-23 & \checkmark&\checkmark &95.99&90.12&51.23&26.97&66.08\\
FLoR-FT&ViT/DINO &  CVPR-24 & \checkmark&\checkmark &96.47&90.75&53.06&27.02&66.83\\
DAMIM-FT&ViT/DINO & AAAI-25 & \checkmark&\checkmark &96.34&91.18&54.86&27.82&67.78\\
IM-DCL & RN10 &  TIP-24 & -&\checkmark &95.73&89.47&52.74&\textbf{28.93}&66.72\\
StepSPT&ViT/CLIP & TPAMI-25 & -&\checkmark &96.01&89.40&52.12&26.36&65.97\\
SeGD-VPT&ViT/CLIP & MM-24 & -&\checkmark &96.93&93.81&53.10&23.20&66.76\\
\midrule 
CLIP-LoRA & ViT/CLIP & CVPR-24 & - & \checkmark & 96.20 & 92.63 & 50.68 & 24.44 & 65.99 \\  \rowcolor{cyan!10}
CLIP-LoRA + \textbf{Ours}&ViT/CLIP & Ours & -&\checkmark  & \textbf{97.08} & \textbf{94.35} & \textbf{54.72} & 25.47 & \textbf{67.90} \\
\textcolor{teal}{\small\textbf{$\Delta$}} & - & - & - & - &
\textcolor{teal}{\small\textbf{+0.88}} & \textcolor{teal}{\small\textbf{+1.72}} &
\textcolor{teal}{\small\textbf{+4.04}} & \textcolor{teal}{\small\textbf{+1.03}} &
\textcolor{teal}{\small\textbf{+1.91}} \\
\bottomrule
\end{tabular}
}  
\label{tab:5shot} 
\vspace{-0.15cm}
\end{table*}

\begin{table*}[th]
\setlength{\abovecaptionskip}{2pt}  
\caption{Comparison with state-of-the-art CDFSL works by the 5-way 1-shot classification.}
\centering
\resizebox{0.9\textwidth}{!}{
\begin{tabular}{lcc*{7}{c}} 
\toprule
Method & Backbone & Mark & Source & Target & CropDiseases & EuroSAT & ISIC & ChestX & Ave.\\
\midrule
MEM-FS&ViT/DINO & TIP-23 & & &81.11&68.11&32.97&22.76&51.24\\
StyleAdv&ViT/DINO & CVPR-23 &\checkmark&- &81.22&72.15&33.05&22.92&52.34\\
FLoR&ViT/DINO & CVPR-24 & \checkmark&- &81.81&72.39&34.20&22.78&52.80\\
DAMIM&ViT/DINO & AAAI-25  & \checkmark&- &82.34&72.87&34.66&22.97&53.21\\
AttnTemp&ViT/DINO & NeurIPS-24 & \checkmark&- &84.02&74.35&34.92&23.19&54.12\\
CD-CLS&ViT/DINO & NeurIPS-24 & \checkmark&\checkmark &84.53&74.97&35.56&23.39&54.62\\
PMF&ViT/DINO & CVPR-22 & \checkmark&\checkmark &80.79&70.74&30.36&21.73&50.91\\
StyleAdv-FT&ViT/DINO &  CVPR-23 & \checkmark&\checkmark &84.11&74.93&33.99&22.92&53.99\\
FLoR-FT&ViT/DINO &  CVPR-24 & \checkmark&\checkmark &83.55&73.09&35.49&23.26&53.85\\
DAMIM-FT&ViT/DINO & AAAI-25 & \checkmark&\checkmark &83.90&73.61&36.35&23.38&54.31\\
IM-DCL&RN10 & TIP-24 & -& \checkmark &84.37&77.14&38.13&\textbf{23.98}&55.91\\
StepSPT&ViT/CLIP & TPAMI-25 & -& \checkmark &84.84&70.01&32.97&22.84&52.68\\
\midrule 
CLIP-LoRA & ViT/CLIP & CVPR-24 & - & \checkmark & 85.11  & 81.49 & 35.23 & 21.73 & 55.89 \\  \rowcolor{cyan!10}
CLIP-LoRA +\textbf{Ours}&ViT/CLIP & Ours & - & \checkmark & \textbf{88.91} &\textbf{86.07} & \textbf{38.13} &22.21&\textbf{58.83}\\
\textcolor{teal}{\small		\textbf{$\Delta$}} & -	& - & - & - &	\textcolor{teal}{\small	\textbf{+3.80}} & 	\textcolor{teal}{\small	\textbf{+4.58}} & 		\textcolor{teal}{\small		\textbf{+2.90}} & 		\textcolor{teal}{\small		\textbf{+0.48}} & 		\textcolor{teal}{\small		\textbf{+2.94}} \\
\bottomrule
\end{tabular}
}  
\label{tab:1shot}
\vspace{-0.35cm}
\end{table*}

\section{Interpretability}
\textbf{T-I-T cycle pathway}
Figure \ref{t2tt} presents additional visualizations of the Text-to-Image-to-Text (T-I-T) cycle consistency pathway. The pathways demonstrate the model's ability to establish fine-grained semantic connections between textual descriptions and visual regions. Even when the reconstructed text does not exactly match the original text  (e.g., ``River" vs. ``Highway or Road"  of the EuroSAT for Pathway-2), the inconsistency provides valuable insights into the model's understanding and reasoning at a local level. For instance, both ``River" and ``Highway or Road"  categories represent linear  structures (elongated shapes) in satellite imagery. 
Note that while the Semantic Anchor (SA) Module's augmentation phase expands the corpus for the first hop of the T-I-T cycle, for display convenience, only the original image is used for retrieval in these visualizations. 

\noindent \textbf{I-T-I cycle pathway}
As explicitly mentioned in the main paper, Figure \ref{i2ii} provides more detailed visualizations 
of the Image-to-Text-to-Image (I-T-I) cycle consistency pathway: the model (1) extracts semantics 
from an anchor patch (red box), (2) maps it to the most similar text label, and (3) retrieves a 
semantically matching patch only in the augmented view (red box). 
It highlights the model's ability to maintain semantic consistency across different visual 
transformations and through the textual modality.

\section{Generalization beyond CLIP}
We focus on  the Source-Free Cross-Domain Few-Shot Learning (SF-CDFSL) benchmark  
established by StepSPT~\cite{DBLP:journals/corr/abs-2411-10070} 
and SeGD-VPT~\cite{DBLP:journals/corr/abs-2412-00767},   
where only a pre-trained model and scarce target-domain shots are available; CLIP is 
widely adopted as the default backbone in this setting for fair comparison.  
To verify our findings and method are generalizable to other baselines, we take SigLIP2~\cite{tschannen2025siglip2multilingualvisionlanguage} and PE-Core~\cite{bolya2025perceptionencoderbestvisual} 
as two vision-language baselines with strengthened fine-grained representations. 
Specifically, SigLIP2 integrates  Location-aware Captioners (LocCa)  and   vision-only  
self-supervised learning 
(including SILC and TIPS)   
to enhance dense prediction and localization capabilities.   
PE-Core aligns and tunes intermediate-layer features  
to 
capture fine-grained spatial  representations. 
As presented in Tab.~\ref{backbone}, our method achieves consistent performance gains across diverse backbones.

\begin{table}[H] 
\vspace{-0.25cm}  
\setlength{\abovecaptionskip}{2pt}  
\caption{Accuracy  
on different backbones in 1-shot.} 
\centering 
\small 
\setlength{\tabcolsep}{1mm} 
\resizebox{0.48\textwidth}{!}{    
\begin{tabular}{lcc*{5}{c}} 
\toprule 
Method & Backbone &  ISIC & ChestX & EuroSAT &   
CropDiseases 
 & 
 Average  
 \\
\midrule 
 CLIP-LoRA &    RN50/CLIP &   32.01 & 21.76 & 57.79  & 65.24 & 44.20 \\ 
+ \textbf{OURS} 
&  RN50/CLIP &   \textbf{35.21} & \textbf{22.75} & \textbf{59.23} & \textbf{72.85} & \textbf{47.51} \\ 
\hline  
 SigLIP2-LoRA&    ViT/SigLip2 &    26.48 & 20.53 & 63.05 & 81.84 &  47.98 
 \\ 
 + \textbf{OURS} 
&  ViT/SigLip2 &   \textbf{29.53}   & \textbf{22.00} & \textbf{68.12} & \textbf{83.39} & \textbf{50.76} 
\\  
\hline 
PE-Core-LoRA &     ViT/PE-Core &  38.05  & 22.45  & 82.16 & 89.01 & 57.92  \\ 
 + \textbf{OURS} 
&   ViT/PE-Core &  \textbf{40.72} & \textbf{22.67} & \textbf{83.92} & \textbf{90.48} & \textbf{59.45} \\ 
\bottomrule
\end{tabular}
} 
\label{backbone} 
\vspace{-0.3cm} 
\end{table}

Moreover, we also reproduce the alignment experiments (Fig.~\ref{alignment score} in the paper) in Tab.~\ref{tab:local}. 
We can see that although these two methods strengthen the fine-grained representations, the 
phenomenon  of degraded local alignment still exists  under extreme domain shifts. 
Quantitatively, Tab.~\ref{tab:local} shows that our method consistently improves local alignment 
scores by \textbf{+26.3\%} for SigLIP2 and \textbf{+18.6\%} for PE-Core, despite different ways to 
pretrain the VLM.


\begin{table}[htbp]
\setlength{\abovecaptionskip}{2pt}  
\caption{Feature alignments of other baselines.}
\centering
\small
\setlength{\tabcolsep}{1mm} 
\resizebox{0.47\textwidth}{!}{    
\begin{tabular}{lcc*{5}{c}}
\toprule
Method & Feature type  &  ISIC & ChestX & EuroSAT &  
CropDiseases 
 & 
 Average 
 \\
\midrule
 SigLIP2-LoRA&    global &    0.1511 & 0.1380 & 0.1226 & 0.1691 &  0.14520 \\ 
 SigLIP2-LoRA&    local  &    0.0330 & 0.0310 & 0.0126 & 0.0383 &  0.02872 \\ 
 + OURS
&   local  &   0.0421   & 0.0363 & 0.0181 &  0.0486 & 0.03627 \\ \hline
PE-Core-LoRA &       global  &   0.2795  & 0.2493  & 0.2459 & 0.2810 & 0.26392 \\  
PE-Core-LoRA &      local  &   0.0191  &  0.0096  & 0.0264 & 0.0088 & 0.01597 \\  
 + OURS 
&    local &  0.0238 & 0.0111 & 0.0286 & 0.0123 & 0.01895 \\ 
\bottomrule
\end{tabular}
}
\label{tab:local}
\vspace{-0.2cm}
\end{table}

In summary, these experiments verify that our findings and designs are generalizable to other 
baselines.

\section{Hyperparameters in Eq.~\ref{eq:loss}}
Table~\ref{tab:hyper} lists the cycle-consistency weights $\lambda_1$ (T-I-T) and $\lambda_2$ (I-T-
I) for each target dataset, determined by grid search on the validation split. All experiments use 
$k=10$ for selecting anchor patches. 

\begin{table}[htbp] 
\vspace{-0.2cm} 
\setlength{\abovecaptionskip}{2pt}
\caption{Hyperparameters $\lambda_1$ and $\lambda_2$ for each dataset.}
\centering
\small
\setlength{\tabcolsep}{1.1mm} 
\resizebox{0.42\textwidth}{!}{    
\begin{tabular}{lccccc}
    \toprule
      & ChestX & ISIC2018 & EuroSAT & CropDiseases \\
    \midrule
    $\lambda_1$  & 3 & 3 & 1.5 & 1 \\
    $\lambda_2$  & 0.5 & 2 & 0.2 & 1.5 \\
    \bottomrule
\end{tabular}
}      
\label{tab:hyper}
\vspace{-0.4cm}
\end{table} 